\documentclass{article}
\usepackage{iclr2027_conference,times}
\usepackage[table]{xcolor}
\usepackage{hyperref}
\usepackage{amsmath}
\usepackage[capitalize,noabbrev]{cleveref}
\usepackage{url}
\usepackage{graphicx}
\usepackage{wrapfig}
\usepackage{needspace}
\usepackage{subcaption}
\usepackage{multirow}
\usepackage{booktabs}
\usepackage{array}
\usepackage{amssymb}
\usepackage{comment}

\newcommand{\ours}{MoLR}
\newcommand{\cmark}{\ensuremath{\checkmark}}
\newcommand{\xmark}{\ensuremath{\times}}
\newcommand{\tablestretch}{1.05}
\newcommand{\best}[1]{\textbf{#1}}
\definecolor{oursrowcolor}{HTML}{96CCCB}
\newcommand{\ourscell}{\cellcolor{oursrowcolor!15}}

\crefname{section}{Sec.}{Secs.}
\Crefname{section}{Sec.}{Secs.}
\crefname{equation}{Eq.}{Eqs.}
\Crefname{equation}{Eq.}{Eqs.}
\crefname{figure}{Fig.}{Figs.}
\Crefname{figure}{Fig.}{Figs.}
\crefname{table}{Tab.}{Tabs.}
\Crefname{table}{Tab.}{Tabs.}
\crefname{appendix}{App.}{Apps.}
\Crefname{appendix}{App.}{Apps.}
\crefname{theorem}{Prop.}{Props.}
\Crefname{theorem}{Prop.}{Props.}
\crefname{algorithm}{Alg.}{Algs.}
\Crefname{algorithm}{Alg.}{Algs.}
\AtBeginDocument{}

\usepackage{amsmath,amsfonts,bm}

\def\eqref#1{equation~\ref{#1}}

\def\1{\bm{1}}

\DeclareMathAlphabet{\mathsfit}{\encodingdefault}{\sfdefault}{m}{sl}
\SetMathAlphabet{\mathsfit}{bold}{\encodingdefault}{\sfdefault}{bx}{n}

\def\sR{{\mathbb{R}}}

\iclrfinalcopy

\title{Mixture of Latent Recursions for Low-Rank Adaptation}
\author{%
  Weixun Luo\qquad Ye Mao\qquad Ranran Huang\qquad
  Junpeng Jing\thanks{Corresponding author (j.jing23@imperial.ac.uk).}\qquad Krystian Mikolajczyk \\
  Imperial College London
}
\begin{document}
\maketitle
\lhead{Preprint}
\vspace{-10pt}
\begin{abstract}
Low-Rank Adaptation (LoRA) performs task-specific adaptation using a fixed low-rank architecture, whose expressiveness is typically increased by enlarging or adding learnable modules with a proportional parameter overhead. We introduce Mixture of Latent Recursions (MoLR), an adaptive recursion framework that enhances adaptation expressiveness while preserving parameter efficiency. MoLR recursively refines the down-projected latent representation of each input using a shared transformation. A lightweight threshold-based router independently determines whether each token should undergo further refinement, while dynamically adjusted thresholds encourage balanced utilization across recursion depths. Extensive experiments across diverse language and vision tasks demonstrate that MoLR consistently outperforms representative LoRA variants with negligible parameter overhead over LoRA. Further analyses show that MoLR learns input-dependent recursion-depth allocation and enables a controllable inference-time compute–performance trade-off without requiring retraining.
\end{abstract}

\section{Introduction}
As foundation models have grown to billions of parameters, full-parameter fine-tuning has become increasingly resource-intensive~\citep{zhang2023adalora}. To mitigate this challenge, Low-Rank Adaptation (LoRA)~\citep{hu2022lora} has emerged as a prominent parameter-efficient fine-tuning mechanism owing to its simple formulation and broad applicability. Rather than directly updating the pre-trained model weights, LoRA keeps them frozen and parameterizes task-specific weight updates as the product of two trainable low-rank matrices. By optimizing these lightweight matrices only, LoRA can achieve performance comparable to, and in some cases exceeding, that of full-parameter fine-tuning while reducing the associated computational and memory costs.

Beyond enabling efficient fine-tuning, the low-rank factorization underlying LoRA and many of its subsequent variants~\citep{kalajdzievski2023rslora,zhang2023lorafa,hayou2024loraplus,meng2024pissa,wang2025lorapro} can be naturally interpreted from a latent-space perspective. Under this view, each input is first down-projected into a low-dimensional latent state and then directly mapped to a high-dimensional output update without any intervening processing. Since a single latent entry can affect multiple output dimensions through the learned latent-to-output mapping, transformations confined to this compact space can induce broad and coordinated changes in the overall adaptation. At the same time, the low dimensionality of the latent space offers a natural means of controlling the overhead of such transformations. Building upon these properties, recent work~\citep{balazy2024loraxs,wu2024moslora,dong2025aurora,li2025toplora} has begun to explore introducing explicit latent-space computation to enhance the expressiveness of low-rank adaptation.

However, despite their demonstrated effectiveness, existing approaches that incorporate latent-space computation still struggle to decouple further gains in adaptation expressiveness from corresponding increases in the number of trainable parameters. In these methods, latent transformations are implemented through fixed feed-forward architectures whose learnable modules are separately parameterized. Consequently, realizing more complex latent transformations (e.g., by enlarging the latent dimensionality, stacking additional transformations, or introducing further learnable modules) typically requires expanding the architecture, resulting in parameter overhead that grows with the size or number of its parameterized components. This raises a central question: \textit{how can we enrich latent-space computation without proportional parameter growth?}

In this paper, we propose Mixture of Latent Recursions (\ours{}), an adaptive recursion framework for low-rank adaptation. To answer this question, MoLR recursively refines the latent state of each input through a shared transformation, enabling richer latent-space computation without trainable-parameter growth proportional to the maximum recursion depth. To account for inherent semantic and contextual variation across inputs, we introduce a lightweight threshold-based router that independently determines at each step whether recursion should continue. This realizes input-dependent recursion depths while avoiding the future-token information leakage from the conventional sequence-wide expert-choice routing~\citep{bae2025mor}. To mitigate routing decisions from collapsing to a narrow range of depths, we dynamically adjust the routing thresholds to encourage balanced depth allocation without incurring potentially interfering auxiliary-loss gradients.

We conduct extensive experiments across diverse language and vision tasks. MoLR achieves the highest performance among representative LoRA variants while requiring a parameter budget comparable to LoRA. Our analyses reveal token-wise recursion-depth patterns, suggesting that the learned routing is sensitive to semantic variation across tokens. We also observe test-time scaling within the trained recursion horizon, where larger inference-time depth caps improve performance without retraining. Finally, incorporating adaptive latent recursion into existing LoRA variants consistently improves their performance, demonstrating compatibility across distinct formulations and suggesting latent-space computation as a promising scaling axis for low-rank adaptation.

Our main contributions are summarized as follows:
\begin{itemize}
    \item We propose \ours{}, an adaptive recursion framework for low-rank adaptation.
    \item We introduce adaptive latent recursion, enabling input-dependent allocation of latent-space computation without parameter cost proportional to the maximum recursion depth.
    \item We demonstrate that \ours{} surpasses representative LoRA variants on diverse tasks with token-wise depth allocation. We also showcase its test-time scaling behavior and the compatibility of adaptive latent recursion with distinct existing LoRA formulations.
\end{itemize}

\section{Related Work}
\textbf{Low-Rank Adaptation.}
LoRA~\citep{hu2022lora} enables parameter-efficient adaptation by freezing the pre-trained model parameters and learning low-rank weight updates. These updates can be merged into the base weights after fine-tuning, thereby introducing no additional inference-time computation. Recent work has progressed along three primary directions: (i) improving empirical performance and reducing the gap relative to full-parameter fine-tuning~\citep{ye2026,zhang2026scalora,singhal2026abba}; (ii) characterizing and improving its optimization dynamics, including conditioning, initialization, and the stability of feature learning~\citep{castin2026balora,zhang2026lorada,wu2026stablelora}; and (iii) extending its applicability to scalable multi-task and continual learning settings~\citep{tian2026mtlora,he2026loda,lu2026little}. These advances reinforce LoRA as a flexible and efficient framework for adapting large foundation models.

\textbf{Latent-Space Computation in LoRA.}
To enhance the expressiveness of LoRA, recent work has introduced explicit learnable computation between its down- and up-projections, which can be broadly categorized into: (i) linear transformations~\citep{balazy2024loraxs,wu2024moslora}; (ii) nonlinear transformations~\citep{dong2025aurora,li2025afalora,chen2026cera} and (iii) input-conditioned transformations that adapt the latent operator to individual inputs~\citep{li2025toplora}. These methods rely on fixed feed-forward architectures, such that realizing more complex latent transformations typically requires enlarging that architecture with a corresponding increase in the parameter count. This motivates us to explore how to enrich latent-space computation without proportional parameter growth.

\textbf{Recurrence and Conditional Computation.}
Recurrence was introduced in early neural network models as a parameter-efficient mechanism for increasing computational depth through repeated application of shared transformations~\citep{elman1990finding,hochreiter1997long}. It was later extended to Transformer architectures, where recurrent or weight-tied blocks enable deeper computation, compact parameterizations, and iterative latent-space reasoning~\citep{dehghani2019universal,giannou2023looped,bae2025relaxed,geiping2025recurrent}. More recent work further incorporates conditional depth, allowing computation to vary across inputs or tokens according to their refinement needs~\citep{schuster2022calm,bae2025mor,jeddi2026loopformer}.

\section{Method}
\begin{figure*}[t]
    \centering
    \includegraphics[width=\linewidth]{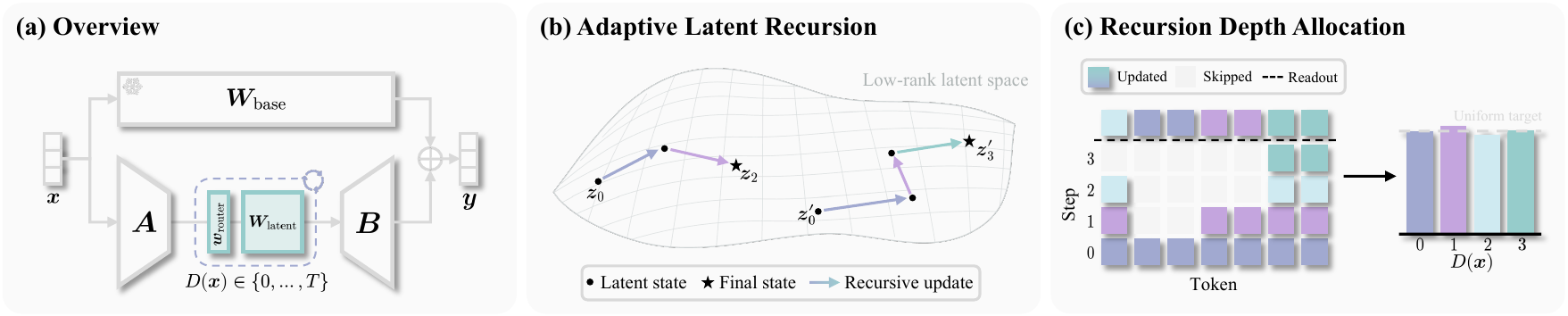}
    \vspace{-15pt}
    \caption{\textbf{\ours{}}. (a) \ours{} recursively refines each latent state to a router-determined depth $D(\boldsymbol{x})$. (b) By using the shared latent transformation, \ours{} yields input-specific trajectories within the latent space without incurring trainable parameter growth as the maximum recursion depth $T$ increases. (c) Once a token halts, subsequent recursive updates are skipped, and its current latent state is retained for readout. We encourage a uniform distribution over stopping depths to promote balanced computation allocation and supervision throughout the recursion horizon.}
    \vspace{-5pt}
    \label{figure:ours}
\end{figure*}

In this section, we first present a latent-space interpretation of LoRA and then introduce \ours{}, an adaptive recursion framework for low-rank adaptation, as shown in~\cref{figure:ours}.


\subsection{Low-Rank Adaptation}
LoRA~\citep{hu2022lora} parameterizes updates to a frozen base weight $\boldsymbol{W}_{\text{base}} \in \mathbb{R}^{d_{\text{out}} \times d_{\text{in}}}$ into learnable low-rank matrices. Given an input $\boldsymbol{x} \in \sR^{d_{\text{in}}}$, the output $\boldsymbol{y}_{\mathrm{LoRA}}\in\mathbb{R}^{d_{\mathrm{out}}}$ is defined as:
\begin{equation}
   \boldsymbol{y}_{\text{LoRA}} = \boldsymbol{W}_{\text{base}}\boldsymbol{x} + \frac{\alpha}{r}\boldsymbol{B}\boldsymbol{z},
   \quad
   \boldsymbol{z} = \boldsymbol{Ax} \in \mathbb{R}^{r},
\end{equation}
where $\boldsymbol{A} \in \mathbb{R}^{r \times d_{\text{in}}}$ and $\boldsymbol{B} \in \mathbb{R}^{d_{\text{out}} \times r}$ denote the down- and up-projection matrices, respectively, $r$ is the rank that is chosen to satisfy $r \ll \min(d_{\text{in}}, d_{\text{out}})$, and $\alpha$ is a scaling coefficient. This formulation exposes a latent-space interpretation of LoRA: Each input $\boldsymbol{x}$ is first encoded by $\boldsymbol{A}$ into an $r$-dimensional latent state $\boldsymbol{z}$, which is directly mapped by $\boldsymbol{B}$ to an output residual without any intervening transformation. A natural way to increase the expressive capacity of LoRA is to enlarge $r$, at the cost of trainable-parameter overhead, $r(d_{\mathrm{in}}+d_{\mathrm{out}})$, that grows proportionally with $r$.

\subsection{Mixture of Latent Recursions}
To decouple the complexity of latent-space computation from proportional growth in trainable parameters, we introduce \ours{}, which recursively refines the latent state through a parameter-shared transformation across recursion steps. Given an input $\boldsymbol{x}$, the output $\boldsymbol{y}_{\text{\ours{}}} \in \mathbb{R}^{d_{\text{out}}}$ is defined as:
\begin{equation}
\boldsymbol{y}_{\text{\ours{}}}
= \boldsymbol{W}_{\text{base}}\boldsymbol{x}
+ \frac{\alpha}{r}\boldsymbol{B}\boldsymbol{z}_{T},
\quad
\boldsymbol{z}_{0}=\boldsymbol{Ax},
\end{equation}
where $T$ denotes the maximum recursion depth and $\boldsymbol{z}_{T} \in \mathbb{R}^r$ is the resulting latent state. For an input with zero executed recursion steps, $\boldsymbol{z}_T=\boldsymbol{z}_0=\boldsymbol{A}\boldsymbol{x}$, so its output coincides with that of LoRA.

\textbf{Adaptive Latent Recursion.}
To account for semantic and contextual variations across inputs, we introduce adaptive latent recursion with state-dependent routing. At each step $t\in\{0,\ldots,T-1\}$, \ours{} first computes a continuation score $s_t\in(0,1)$ for each active input with latent state $\boldsymbol{z}_t$:
\begin{equation}
    \label{equation:routing}
    s_{t}
    = \operatorname{sigmoid}\!\left(
        \boldsymbol{w}_{\text{router}}^{\top}\boldsymbol{z}_{t}
    \right),
\end{equation}
where $\boldsymbol{w}_{\text{router}}\in\mathbb{R}^{r}$ is the shared router weight across steps. The latent state is updated as follows:
\begin{equation}
    \label{equation:latent_recursion}
    \boldsymbol{z}_{t+1}
    =
    \begin{cases}
        \boldsymbol{z}_{t}
        + s_{t}\boldsymbol{W}_{\text{latent}}\!
        \left(\boldsymbol{z}_{t}+\phi(t)\right),
        & \text{if} \quad s_{t}>b_t, \\
        \boldsymbol{z}_{t} & \text{otherwise},
    \end{cases}
\end{equation}
where $\boldsymbol{W}_{\text{latent}}\in\mathbb{R}^{r\times r}$ denotes the shared latent transformation across steps, $b_t \in \mathbb{R}$ is the routing threshold at step $t$, and $\phi:\{0,\ldots,T-1\}\rightarrow\mathbb{R}^{r}$ represents a fixed step-encoding function:
\begin{equation}
    \phi(t)_{2i}
    = \sin\!\left(\frac{2\pi t(i+1)}{T}\right),
    \quad
    \phi(t)_{2i+1}
    = \cos\!\left(\frac{2\pi t(i+1)}{T}\right),
\end{equation}
with $i\in\{0,\ldots,r/2-1\}$ indexing coordinate pairs across $r$ latent dimensions. Herein, an active input undergoes further refinement only when its state-dependent continuation score satisfies $s_t>b_t$. For inputs that continue, the step encoding $\phi(t)$ is added to provide explicit recursion-stage information before the shared transformation $\boldsymbol{W}_{\text{latent}}$ is applied. The resulting residual update is scaled by $s_t$, allowing the router to modulate the refinement and receive gradients through the executed update. Once an input fails the continuation criterion, its recursion terminates permanently, and its current latent state is preserved as the final representation $\boldsymbol{z}_{T}$.

For an input $\boldsymbol{x}$ that realizes a recursion depth $D(\boldsymbol{x}) \in \{0,\ldots,T\}$, expanding the recursive update in~\cref{equation:latent_recursion} while omitting $\phi(t)$ for notational clarity yields:
\begin{equation}
\boldsymbol{z}_{D(\boldsymbol{x})}
=
\prod_{t=0}^{D(\boldsymbol{x})-1}
\!\left(
\boldsymbol{I}_r+
s_t\boldsymbol{W}_{\text{latent}}
\right)
\boldsymbol{z}_{0}
=
\sum_{d=0}^{D(\boldsymbol{x})}
\gamma_{d}(\boldsymbol{x})
\boldsymbol{W}_{\text{latent}}^{d}
\boldsymbol{z}_{0},
\end{equation}
where $\gamma_{d}(\boldsymbol{x})$ is determined by the continuation scores $s_t$ across the executed steps, with $\gamma_{0}(\boldsymbol{x})=1$. The expansion shows that adaptive latent recursion induces an input-dependent polynomial in the shared latent transformation $\boldsymbol{W}_{\text{latent}}$. The realized depth $D(\boldsymbol{x})$ determines the maximum transformation order, while $s_t$ governs the contribution of each order. As a result, \ours{} yields a family of effective latent transformations without trainable parameter growth proportional to $T$. Detailed discussion on the exact expansion of~\cref{equation:latent_recursion} is provided in~\cref{appendix:discussion}.

\textbf{Load Balancing.}
Unconstrained routing can concentrate training examples at a narrow range of stopping depths, leaving final adapter outputs at other depths with less direct supervision and potentially limiting the effectiveness of the shared latent transformation when recursion terminates at those depths. To address this issue, we specify a uniform target distribution over the realized recursion depths $D(\boldsymbol{x})$ that is defined as:
\begin{equation}
    \label{equation:uniform_depth_distribution}
    \mathbb{P}_{\text{target}}(D(\boldsymbol{x})=d)=\frac{1}{T+1},
    \quad
    \mathbb{E}_{\text{target}}[D(\boldsymbol{x})] = \frac{T}{2},
    \quad
    \forall d\in\{0,\ldots,T\}.
\end{equation}
\begin{table*}[t]
    \centering
    \caption{\textbf{Structural comparison of LoRA variants.} For each variant, we report its equivalent latent-space operator, additional trainable parameters, and worst-case per-token multiply--accumulate operations (MACs) relative to LoRA. In the latent space, LoRA passes the latent state directly to up-projection; MoSLoRA applies a learnable linear transformation $\boldsymbol{W}_{\text{latent}} \in \mathbb{R}^{r \times r}$; TopLoRA applies an input-dependent diagonal transformation $\boldsymbol{\Sigma}(\boldsymbol{x})\in\mathbb{R}^{r\times r}$; and \ours{} performs  adaptive recursive refinement with input-dependent depths.}

    \setlength{\tabcolsep}{2.5pt}
    \begin{tabular*}{\textwidth}{@{\extracolsep{\fill}}lccccc@{}}
        \toprule
        \textbf{Method}
        & \textbf{Input-Conditioned}
        & \textbf{Adaptive Depth}
        & \textbf{Latent Operator}
        & $\boldsymbol{\Delta} $\textbf{Params.}
        & $\boldsymbol{\Delta} $\textbf{MACs} \\
        \midrule
        LoRA
        & \xmark
        & \xmark
        & $\boldsymbol{I}_r$
        & --
        & -- \\
        MoSLoRA
        & \xmark
        & \xmark
        & $\boldsymbol{W}_{\text{latent}}$
        & $r^2$
        & $r^2$ \\
        TopLoRA
        & \cmark
        & \xmark
        & $\boldsymbol{\Sigma}(\boldsymbol{x})$
        & $rd_{\text{in}}$
        & $rd_{\text{in}}$ \\
        \ours{}
        & \cmark
        & \cmark
        & $\sum_{d=0}^{D(\boldsymbol{x})}\!\gamma_d(\boldsymbol{x})\boldsymbol{W}_{\text{latent}}^d$
        & $r^2+r$
        & $T(r^2+r)$ \\
        \bottomrule
    \end{tabular*}
    \label{table:structural_comparison}
    \vspace{-5pt}
\end{table*}


The target fraction of active inputs that should undergo another refinement at step $t$ is then:
\begin{equation}
    q_t
    =
    \mathbb{P}_{\text{target}}(D(\boldsymbol{x})\geq t+1\mid D(\boldsymbol{x})\geq t)
    = \frac{\mathbb{P}_{\text{target}}(D(\boldsymbol{x})\geq t+1)}
         {\mathbb{P}_{\text{target}}(D(\boldsymbol{x})\geq t)}
    =
    \frac{T-t}{T-t+1}.
\end{equation}
Following~\citet{wang2024lossfree}, we update the threshold at each recursion step using:
\begin{equation}
    \label{equation:load_balancing}
    b_t
    \leftarrow
    b_t
    + \eta\operatorname{sign}\!\left(\hat{q}_t-q_t\right),
\end{equation}
where $\eta \in \mathbb{R}_{>0}$ denotes the threshold learning rate, and $\hat{q}_t$ is the observed fraction of active inputs that continue at step $t$. We zero-initialize $b_t$ so that every input initially traverses the full recursion horizon. This avoids premature suppression of recursion due to unreliable early routing decisions, facilitating the emergence of diverse depth allocation patterns in subsequent training.

\textbf{Computational Overhead.}
\cref{table:structural_comparison} summarizes the equivalent latent operators, additional trainable parameters, and worst-case per-token multiply--accumulate operations (MACs) of different LoRA variants. It can be seen that the worst-case operational overhead of \ours{} relative to LoRA is:
\begin{equation}
\frac{T(r^2+r)}
{r(d_{\text{in}}+d_{\text{out}})}
=
\frac{T(r+1)}
{d_{\text{in}}+d_{\text{out}}}.
\end{equation}
Under the standard low-rank regime, where $r\ll\min(d_{\text{in}},d_{\text{out}})$ and the maximum recursion depth $T$ is small, this ratio remains limited. Thus, \ours{} realizes adaptive, input-conditioned higher-order latent transformations while preserving a computational footprint comparable to that of LoRA. Moreover, this expression represents a worst-case bound in which every token executes all $T$ recursion steps. In practice, early stopping allows latent states to terminate at shallower recursion depths, thereby avoiding unnecessary transformations and reducing the realized computational cost.

\section{Experiments}
In this section, we evaluate \ours{} across a broad range of architectures and tasks, covering natural language understanding (\cref{subsection:natural_language_understanding}), natural language reasoning (\cref{subsection:natural_language_reasoning}), and image classification (\cref{subsection:image_classification}). We then examine its key characteristics in~\cref{subsection:analysis} and~\cref{appendix:analysis} and conclude with ablation studies in~\cref{subsection:ablation_studies} that explore the design choices for individual components.
\begin{table*}[t!]
    \small
    \centering
    \setlength{\fboxsep}{2pt}
    \caption{\textbf{Quantitative comparison of LoRA variants on NLU.} We report accuracy ($\uparrow$) for all tasks except MRPC, QQP, and STSB, for which the combined score ($\uparrow$) is used. \#Param. denotes the number of trainable parameters. The \textbf{best} results are highlighted.}
    \setlength{\tabcolsep}{2.8pt}
    \renewcommand{\arraystretch}{\tablestretch}
    \begin{tabular}{
        c
        l
        w{c}{1.3cm}
        w{c}{1.0cm}
        w{c}{1.0cm}
        w{c}{1.0cm}
        w{c}{1.0cm}
        w{c}{1.0cm}
        w{c}{1.0cm}
        w{c}{1.0cm}
        |w{c}{1.0cm}
    } 
        \toprule
        \textbf{Model}
        & \textbf{Method} 
        & \textbf{\#Param.}
        & \textbf{CoLA} 
        & \textbf{MNLI}
        & \textbf{MRPC} 
        & \textbf{QNLI} 
        & \textbf{QQP} 
        & \textbf{SST-2} 
        & \textbf{STSB} 
        & \textbf{Avg.} 
        \\ 
        \midrule
        \multirow{8}{*}{\rotatebox{90}{RoBERTa-B}}
        & LoRA      & 1.33M & 77.28 & 85.55 & 78.19 & 90.31 & 86.20 & 93.27 & 81.78 & 84.65 \\
        & DoRA      & 1.41M & 77.75 & 85.27 & 76.56 & 90.27 & 86.20 & 93.42 & 81.74 & 84.46 \\
        & AdaLoRA   & 1.33M & 77.95 & 85.44 & 75.96 & 90.94 & 86.60 & 93.46 & 85.39 & 85.11 \\
        & TopLoRA   & 1.99M & 78.84 & 85.24 & 76.50 & 90.40 & 86.08 & 93.39 & 85.62 & 85.15 \\
        & HydraLoRA & 1.58M & 76.38 & 85.27 & 79.46 & 89.97 & 85.88 & 93.27 & 78.56 & 84.11 \\
        & MoSLoRA   & 1.33M & 76.25 & 85.45 & 74.80 & 90.35 & 86.07 & 93.27 & 80.43 & 83.80 \\
        & AuroRA    & 1.34M & 77.37 & 85.57 & 76.27 & 90.43 & 86.16 & 93.35 & 85.40 & 84.94 \\
        & \ourscell\ours{} & \ourscell 1.33M & \ourscell\best{79.86} & \ourscell\best{85.61} & \ourscell\best{82.10} & \ourscell\best{91.04} & \ourscell\best{86.62} & \ourscell\best{93.62} & \ourscell\best{86.73} & \ourscell\best{86.51} \\
        \midrule
        \multirow{8}{*}{\rotatebox{90}{DeBERTa-v3-L}}
        & LoRA      & 3.54M & 86.45 & 91.80 & 74.92 & 95.25 & 89.74 & 95.60 & 86.06 & 88.55 \\
        & DoRA      & 3.76M & 86.64 & 91.68 & 74.92 & 95.21 & 89.69 & 95.68 & 86.44 & 88.61 \\
        & AdaLoRA   & 3.56M & 87.41 & 91.75 & 81.29 & 95.31 & 89.87 & 95.72 & 88.64 & 90.00 \\
        & TopLoRA   & 5.31M & 87.25 & 91.77 & 82.24 & 95.23 & 89.46 & 95.87 & 89.78 & 90.23 \\
        & HydraLoRA & 4.20M & 85.94 & 91.62 & 75.30 & 95.08 & 89.60 & 95.91 & 85.22 & 88.38 \\
        & MoSLoRA   & 3.55M & 86.19 & 91.57 & 77.18 & 95.13 & 89.64 & 96.10 & 85.54 & 88.77 \\
        & AuroRA    & 3.56M & 86.29 & 91.61 & 77.87 & 95.17 & 89.45 & 95.83 & 85.99 & 88.89 \\
        & \ourscell\ours{} & \ourscell 3.55M & \ourscell\best{87.50} & \ourscell\best{91.83} & \ourscell\best{84.96} & \ourscell\best{95.40} & \ourscell\best{90.31} & \ourscell\best{96.29} & \ourscell\best{90.85} & \ourscell\best{91.02} \\
        \bottomrule
    \end{tabular}
    \label{table:natural_language_understanding}
    \vspace{-5pt}
\end{table*}


\textbf{Implementation Details.}
We implement all models in PyTorch and train them on NVIDIA A100 GPUs. Unless otherwise specified, we apply adapters to all linear modules in the backbone, with $r=8$ and $\alpha=2r$ used across all baselines. For each baseline, we adopt the recommended hyperparameters from its paper or official codebase whenever available. For \ours{}, we use $T=2$ and $\eta=1\times10^{-3}$ across all tasks. We repeat every experiment with three different random seeds and report the average performance, with the across-seed variability provided in~\cref{appendix:analysis}.

\textbf{Baselines.}
We compare \ours{} against representative LoRA variants spanning diverse methodological categories, including: (i) LoRA~\citep{hu2022lora}; (ii) DoRA~\citep{liu2024dora}, which decomposes the pre-trained weight into magnitude and direction, learning the magnitude separately while using LoRA to update the direction; (iii) AdaLoRA~\citep{zhang2023adalora}, which adaptively allocates rank budget among weight matrices; (iv) TopLoRA~\citep{li2025toplora}, which enables token-wise input-output projections; (v) HydraLoRA~\citep{tian2024hydralora}, which shares a down-projection across multiple up-projections and dynamically combines their outputs; (vi) MoSLoRA~\citep{wu2024moslora}, which inserts a trainable mixer to fuse subspaces in LoRA; and (vii) AuroRA~\citep{dong2025aurora}, which introduces adaptive nonlinearity between LoRA linear projectors.

\subsection{Natural Language Understanding (NLU)}
\label{subsection:natural_language_understanding}
\textbf{Settings.}
We fine-tune RoBERTa-B~\citep{liu2019roberta} and DeBERTa-v3-L~\citep{he2021debertav3} on seven tasks from the GLUE benchmark~\citep{wang2019glue}. Detailed settings are provided in~\cref{appendix:natural_language_understanding}.

\textbf{Results.}
\cref{table:natural_language_understanding} shows that \ours{} achieves the highest average performance across both backbones and leads on nearly all language understanding tasks. Despite using a parameter budget comparable to LoRA, it outperforms higher-parameter variants such as TopLoRA and methods with explicit nonlinearity like AuroRA. These results highlight the effectiveness of adaptive latent recursion.

\subsection{Natural Language Reasoning (NLR)}
\label{subsection:natural_language_reasoning}
\textbf{Settings.}
We fine-tune Llama-3.2-3B~\citep{grattafiori2025llama} and Qwen3-8B~\citep{yang2025qwen3} on tasks spanning commonsense reasoning, reading comprehension, general knowledge understanding, mathematical reasoning, and code generation. Detailed settings are provided in~\cref{appendix:natural_language_reasoning}.

\textbf{Results.}
\cref{table:natural_language_reasoning} shows that \ours{} achieves the highest average performance on both backbones, obtaining the best results on five tasks with LLaMA-3.2-3B and four tasks with Qwen3-8B. The improvements are particularly pronounced on ARC-C, with consistent gains on BoolQ, GSM8K, and WinoGrande. These results suggest that \ours{} generalizes beyond language understanding, delivering consistent improvements across diverse reasoning tasks, backbones, and model scales.
\begin{table*}[t!]
    \small
    \centering
    \setlength{\fboxsep}{2pt}
    \caption{\textbf{Quantitative comparison of LoRA variants on NLR.}
    We report accuracy ($\uparrow$) for all other tasks and Pass@1 ($\uparrow$) for HumanEval. \#Param. denotes the number of trainable parameters. The \textbf{best} results are highlighted. HS = HellaSwag, HE = HumanEval, WG = WinoGrande.}
    \setlength{\tabcolsep}{2.8pt}
    \renewcommand{\arraystretch}{\tablestretch}
    \begin{tabular}{
        c
        l
        w{c}{1.3cm}
        w{c}{1.0cm}
        w{c}{1.0cm}
        w{c}{1.0cm}
        w{c}{1.0cm}
        w{c}{1.0cm}
        w{c}{1.0cm}
        w{c}{1.0cm}
        |w{c}{1.0cm}
    } 
        \toprule
        \textbf{Model}
        & \textbf{Method} 
        & \textbf{\#Param.}
        & \textbf{ARC-C} 
        & \textbf{BoolQ}
        & \textbf{GSM8K} 
        & \textbf{HS} 
        & \textbf{HE} 
        & \textbf{MMLU} 
        & \textbf{WG} 
        & \textbf{Avg.} 
        \\ 
        \midrule
        \multirow{8}{*}{\rotatebox{90}{LLaMA-3.2-3B}}
        & LoRA      & 12.16M & 49.46 & 88.20 & 34.49 & 78.99 & 23.98 & 55.73 & 83.64 & 59.21 \\
        & DoRA      & 12.93M & 50.11 & 88.16 & 35.03 & 79.02 & 24.19 & 55.67 & 83.80 & 59.42 \\
        & AdaLoRA   & 12.16M & 46.19 & 81.91 & 23.08 & 77.96 & 25.20 & 55.79 & 78.98 & 55.59 \\
        & TopLoRA   & 18.12M & 50.97 & 88.84 & 39.83 & \best{79.42} & 24.19 & \best{56.25} & 83.58 & 60.44 \\
        & HydraLoRA & 14.51M & 47.84 & 87.41 & 33.54 & 78.92 & 23.58 & 55.75 & 83.21 & 58.61 \\
        & MoSLoRA   & 12.17M & 52.16 & 89.33 & 38.06 & 78.78 & 22.36 & 54.46 & 83.71 & 59.84 \\
        & AuroRA    & 12.18M & 52.64 & 88.91 & 35.51 & 79.05 & 24.80 & 53.34 & 83.74 & 59.71 \\
        & \ourscell\ours{} & \ourscell 12.17M & \ourscell\best{54.69} & \ourscell\best{89.52} & \ourscell\best{40.18} & \ourscell 79.11 & \ourscell\best{25.61} & \ourscell 55.48 & \ourscell\best{83.90} & \ourscell\best{61.21} \\
        \midrule
        \multirow{8}{*}{\rotatebox{90}{Qwen-3-8B}}
        & LoRA      & 21.82M & 59.61 & 90.12 & 81.43 & 79.51 & 59.35 & 74.02 & 82.90 & 75.28 \\
        & DoRA      & 23.22M & 60.47 & 90.06 & 82.11 & 79.51 & 59.15 & 74.02 & 82.80 & 75.45 \\
        & AdaLoRA   & 21.82M & 56.91 & 89.61 & 81.58 & 79.50 & 60.98 & 73.83 & 83.40 & 75.11 \\
        & TopLoRA   & 32.44M & 62.88 & 90.72 & 81.35 & \best{79.90} & 55.49 & \best{74.40} & 83.53 & 75.47 \\
        & HydraLoRA & 26.10M & 56.40 & 89.67 & 81.81 & 79.25 & 61.18 & 74.01 & 82.37 & 74.96 \\
        & MoSLoRA   & 21.84M & 57.39 & 89.78 & 81.60 & 79.36 & \best{61.59} & 73.99 & 82.90 & 75.23 \\
        & AuroRA    & 21.86M & 60.24 & 90.85 & 81.53 & 79.19 & 55.90 & 73.61 & 82.85 & 74.88 \\
        & \ourscell\ours{} & \ourscell 21.84M & \ourscell\best{64.53} & \ourscell\best{91.57} & \ourscell\best{82.21} & \ourscell 79.66 & \ourscell 59.15 & \ourscell 74.09 & \ourscell\best{83.66} & \ourscell\best{76.41} \\
        \bottomrule
    \end{tabular}
    \label{table:natural_language_reasoning}
    \vspace{-5pt}
\end{table*}


\subsection{Image Classification (IC)}
\label{subsection:image_classification}

\textbf{Settings.}
We fine-tune CLIP-ViT-B/16~\citep{radford2021clip} and SigLIP-ViT-L/16~\citep{zhai2023siglip} on seven image classification benchmarks. Detailed settings are provided in~\cref{appendix:image_classification}.

\textbf{Results.}
As shown in~\cref{table:image_classification}, \ours{} achieves the highest average accuracy on both backbones and the best performance on every dataset. It consistently outperforms all baselines, with the largest margin over the next-best method observed on DTD. Together with the language results shown in~\cref{table:natural_language_understanding,table:natural_language_reasoning}, these findings indicate that adaptive latent recursion transfers effectively to vision tasks, highlighting its applicability across different architectures and modalities.
\begin{table*}[t!]
    \small
    \centering
    \setlength{\fboxsep}{2pt}
    \caption{\textbf{Quantitative comparison of LoRA variants on IC.} We report the top-$1$ test accuracy ($\uparrow$) for all datasets. \#Param. denotes the number of trainable parameters. The \textbf{best} results are highlighted. Abbreviations: ESAT = EuroSAT, RE45 = RESISC45.}
    \setlength{\tabcolsep}{2.8pt}
    \renewcommand{\arraystretch}{\tablestretch}
    \begin{tabular}{
        c
        l
        w{c}{1.3cm}
        w{c}{1.0cm}
        w{c}{1.0cm}
        w{c}{1.0cm}
        w{c}{1.0cm}
        w{c}{1.0cm}
        w{c}{1.0cm}
        w{c}{1.0cm}
        |w{c}{1cm}
    }
        \toprule
        \textbf{Model}
        & \textbf{Method}
        & \textbf{\#Param.}
        & \textbf{Cars}
        & \textbf{DTD}
        & \textbf{ESAT}
        & \textbf{GTSRB}
        & \textbf{RE45}
        & \textbf{SUN397}
        & \textbf{SVHN}
        & \textbf{Avg.}
        \\
        \midrule
        \multirow{8}{*}{\rotatebox{90}{CLIP-ViT-B}}
        & LoRA      & 1.33M & 71.65 & 52.20 & 97.62 & 97.08 & 91.88 & 73.45 & 96.59 & 82.92 \\
        & DoRA      & 1.41M & 71.85 & 52.38 & 97.67 & 97.15 & 92.06 & 73.56 & 96.62 & 83.04 \\
        & AdaLoRA   & 1.33M & 70.54 & 46.24 & 97.59 & 97.30 & 92.13 & 74.32 & 96.58 & 82.10 \\
        & TopLoRA   & 1.99M & 75.05 & 56.34 & 97.98 & 97.71 & 93.53 & 75.10 & 96.88 & 84.66 \\
        & HydraLoRA & 1.58M & 70.32 & 49.65 & 97.50 & 96.82 & 91.26 & 72.48 & 96.36 & 82.06 \\
        & MoSLoRA   & 1.33M & 70.57 & 50.37 & 97.39 & 96.78 & 90.97 & 72.61 & 96.38 & 82.15 \\
        & AuroRA    & 1.34M & 70.54 & 52.41 & 97.45 & 96.37 & 90.68 & 72.07 & 96.29 & 82.26 \\
        & \ourscell\ours{} & \ourscell 1.33M & \ourscell\best{77.55} & \ourscell\best{65.05} & \ourscell\best{98.17} & \ourscell\best{98.22} & \ourscell\best{94.43} & \ourscell\best{75.90} & \ourscell\best{97.07} & \ourscell\best{86.63} \\
        \midrule
        \multirow{8}{*}{\rotatebox{90}{SigLIP-ViT-L}}
        & LoRA      & 3.54M & 94.91 & 66.13 & 97.94 & 98.23 & 93.41 & 78.42 & 97.13 & 89.45 \\
        & DoRA      & 3.76M & 95.01 & 66.40 & 97.80 & 98.31 & 93.62 & 78.54 & 97.18 & 89.55 \\
        & AdaLoRA   & 3.54M & 94.81 & 64.70 & 97.92 & 98.62 & 94.13 & 80.01 & 97.18 & 89.62 \\
        & TopLoRA   & 5.31M & 95.33 & 68.12 & 98.25 & 98.75 & 94.91 & 80.33 & 97.35 & 90.44 \\
        & HydraLoRA & 4.20M & 94.72 & 65.13 & 97.52 & 97.78 & 92.64 & 77.58 & 96.95 & 88.90 \\
        & MoSLoRA   & 3.55M & 94.78 & 65.39 & 97.59 & 98.08 & 93.03 & 77.99 & 97.00 & 89.12 \\
        & AuroRA    & 3.56M & 94.87 & 67.27 & 97.57 & 97.37 & 92.18 & 77.55 & 96.73 & 89.08 \\
        & \ourscell\ours{} & \ourscell 3.55M & \ourscell\best{95.77} & \ourscell\best{73.30} & \ourscell\best{98.31} & \ourscell\best{98.89} & \ourscell\best{95.39} & \ourscell\best{80.91} & \ourscell\best{97.45} & \ourscell\best{91.43} \\
        \bottomrule
    \end{tabular}
    \label{table:image_classification}
    \vspace{-15pt}
\end{table*}


\subsection{Analysis}
\label{subsection:analysis}

\begin{figure*}[t!]
    \centering
    \captionsetup[subfigure]{aboveskip=-1pt}
    \begin{subfigure}[t]{0.49\textwidth}
        \centering
        \includegraphics[width=\linewidth]{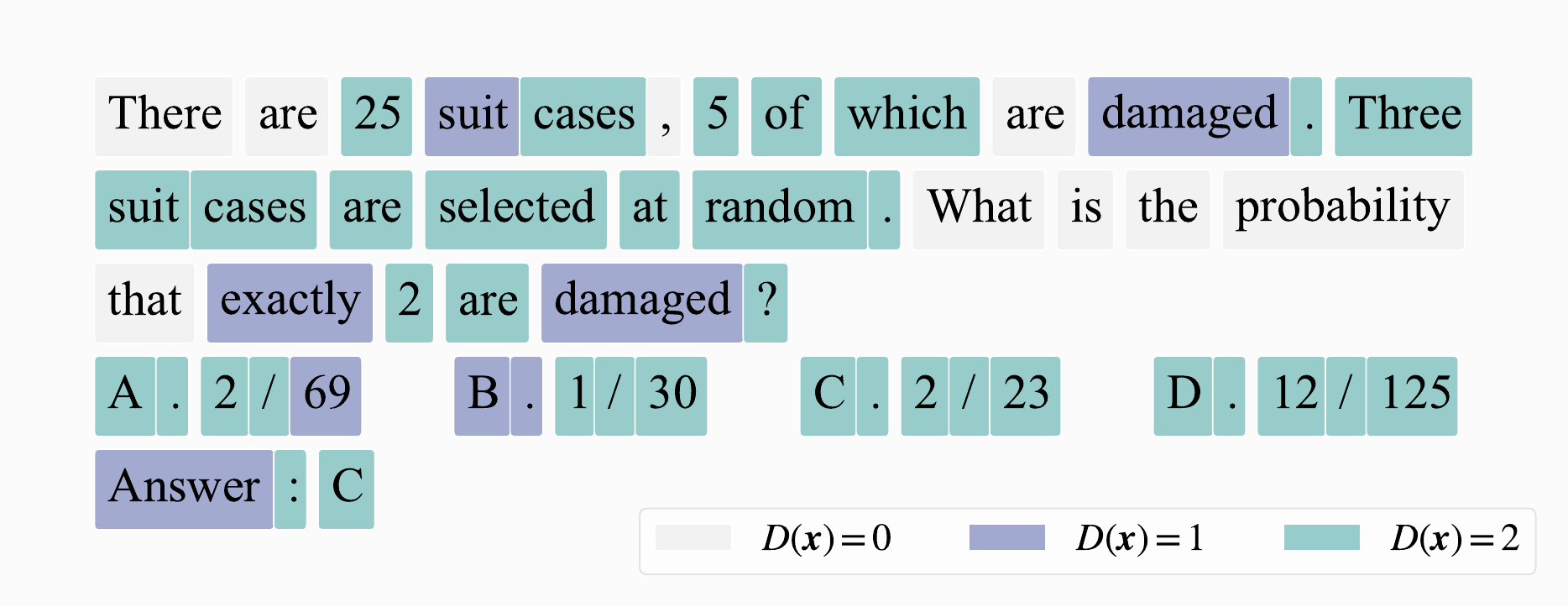}
        \caption{}
        \label{subfigure:rank_per_token}
    \end{subfigure}
    \hfill
    \begin{subfigure}[t]{0.24\textwidth}
        \centering
        \includegraphics[width=\linewidth]{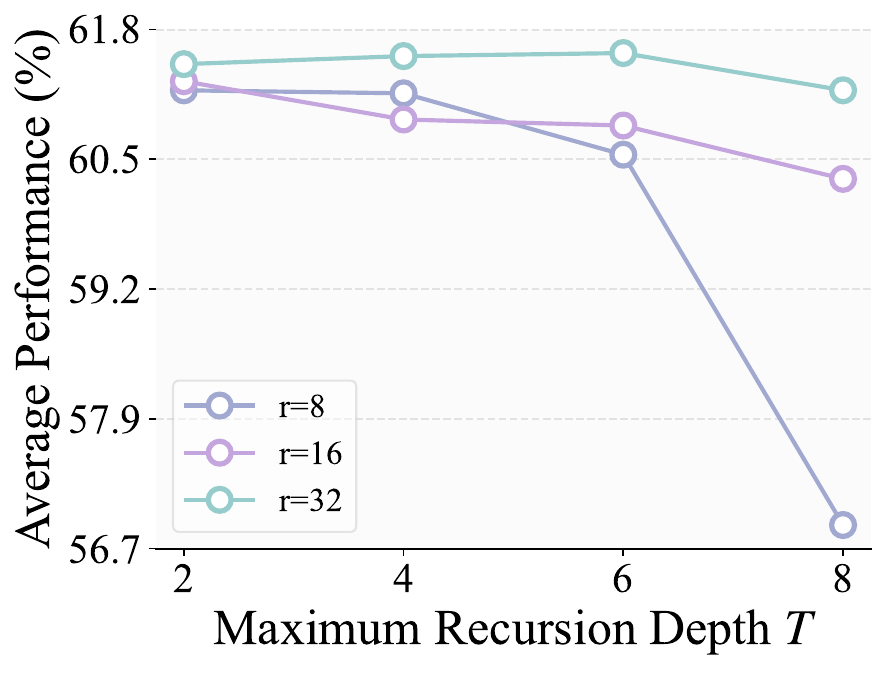}
        \caption{}
        \label{subfigure:recursion_depth_sweep}
    \end{subfigure}
    \hfill
    \begin{subfigure}[t]{0.24\textwidth}
        \centering
        \includegraphics[width=\linewidth]{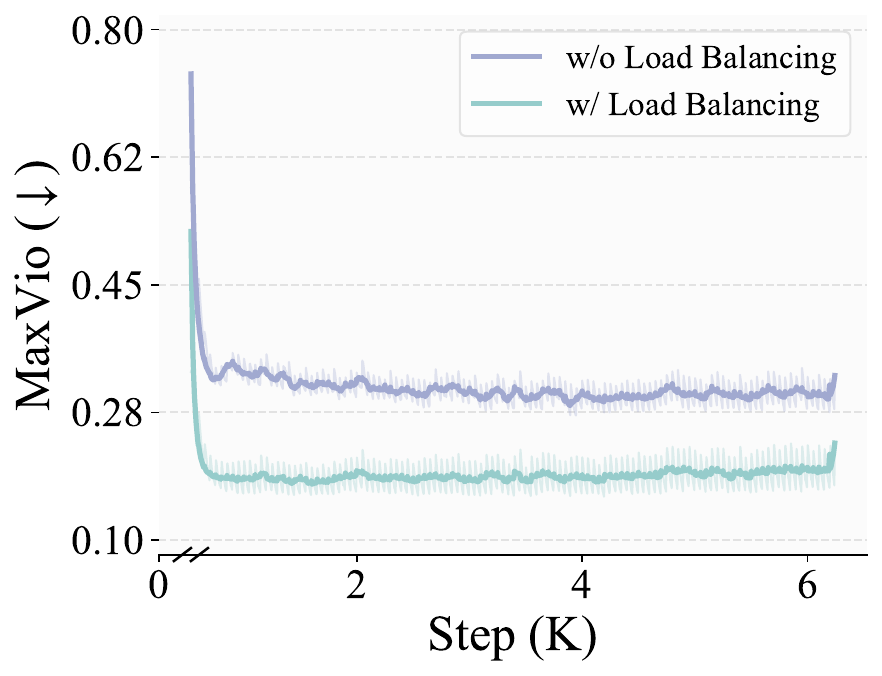}
        \caption{}
        \label{subfigure:load_balancing}
    \end{subfigure}
    \vspace{-10pt}
    \caption{\textbf{Analysis of adaptive latent recursion in \ours{} on NLR.} (a) An example token-wise recursion depth allocation pattern on MMLU when $T = 2$. Each token is colored according to its most frequent recursion depth realized across all modules in layer 8 of Llama-3.2-3B. (b) Average performance when sweeping $T \in \{2,4,6,8\}$ across ranks $r \in \{8,16,32\}$. (c) Comparison of MaxVio ($\downarrow$) with and without load balancing averaged across all modules.}
    \vspace{-10pt}
\end{figure*}

\textbf{Recursion Depth Allocation.}
\cref{subfigure:rank_per_token} visualizes the token-wise recursion depth allocation for an MMLU example. Herein, deeper recursion frequently appears around task-relevant contexts such as ``25 suitcases'' and ``Three are selected,'' whereas question scaffolding like ``There are'' and ``What is'' receives no recursion in most cases. Candidate answers and the final answer are also predominantly processed at maximum recursion depth. This non-uniform depth allocation is consistent with \ours{} learning to adapt latent computation to semantic and contextual differences across tokens.

\textbf{Maximum Recursion Depth.}
\cref{subfigure:recursion_depth_sweep} examines the effect of the maximum recursion depth $T$ across different ranks. At smaller ranks, performance is lower at larger $T$, whereas increasing $r$ substantially attenuates this decline. This interaction suggests that longer recursion horizons may benefit from a more expressive latent space and higher-capacity shared latent transformation, which can better mitigate potential interference induced by reusing the same transformation across recursion steps. At $r=32$, performance saturates and slightly declines at the largest $T$, indicating that an excessively large maximum depth is not inherently beneficial even with greater latent capacity. This observation is consistent with similar saturation or degradation reported in other recursive models~\citep{prairie2026parcae,yang2026stabilizing}. Overall, these results suggest that longer recursion horizons are neither universally beneficial nor necessary once sufficient refinement has been achieved.

\textbf{Recursion Depth Utilization.}
We evaluate how evenly \ours{} utilizes the recursion depths using MaxVio~\citep{wang2024lossfree}, defined as the maximum relative excess over the expected load under uniform depth utilization. We compare against a baseline that resamples the target depth distribution at each step rather than using the fixed distribution defined in~\cref{equation:uniform_depth_distribution}. \cref{subfigure:load_balancing} shows that the threshold update in~\cref{equation:load_balancing} consistently achieves lower MaxVio throughout training, indicating more balanced utilization across recursion depths. It also improves average NLR performance by 1.09 percentage points over the resampling baseline. These results validate the effectiveness of the load-balancing mechanism in promoting balanced optimization across stopping depths.

\textbf{Robustness.}
\cref{sub@subfigure:rank_sweep,sub@subfigure:module_sweep} evaluate \ours{} across varying ranks and target-module coverage. \ours{} shows consistently strong performance across all evaluated configurations. This stability suggests that its effectiveness persists across the evaluated ranks and target-module sets, indicating its robustness across a range of adaptation budgets. Notably, \ours{} with $r=8$ outperforms LoRA with higher ranks despite using fewer trainable parameters. These results suggest that increasing computational expressivity within the latent space through adaptive latent recursion can be more effective and parameter-efficient than simply expanding the latent dimension. Detailed comparisons with additional representative LoRA variants are provided in~\cref{appendix:analysis}.

\begin{figure*}[t!]
    \centering
    \captionsetup[subfigure]{aboveskip=-1pt}
    \begin{subfigure}[t]{0.24\textwidth}
        \centering
        \includegraphics[width=\linewidth]{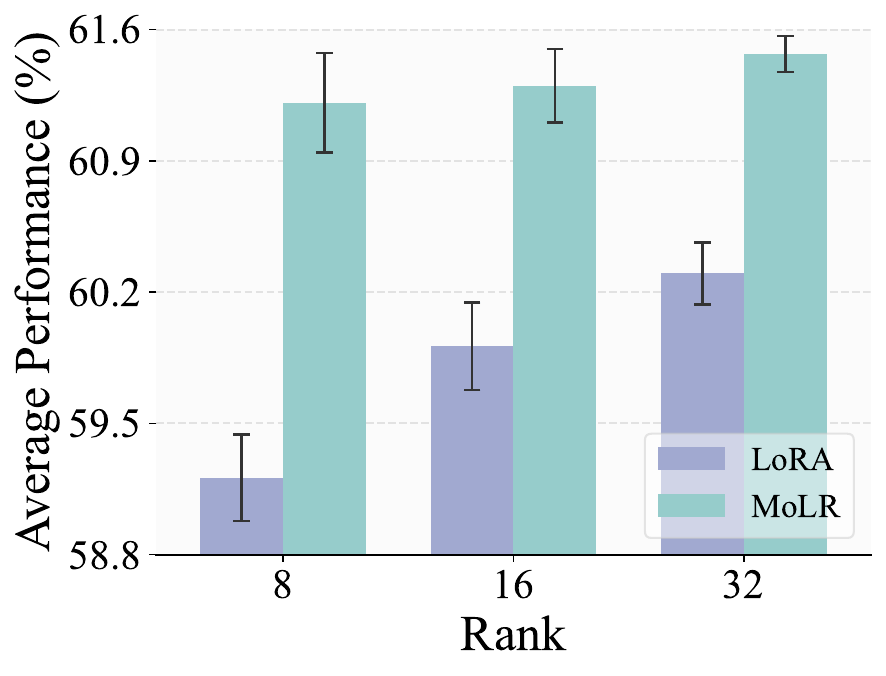}
        \caption{}
        \label{subfigure:rank_sweep}
    \end{subfigure}
    \hfill
    \begin{subfigure}[t]{0.24\textwidth}
        \centering
        \includegraphics[width=\linewidth]{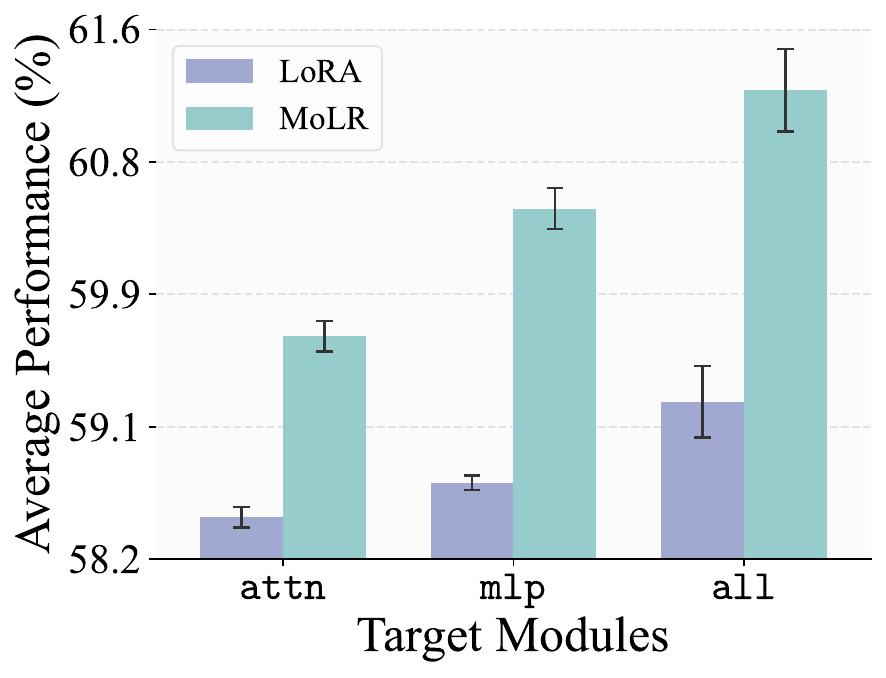}
        \caption{}
        \label{subfigure:module_sweep}
    \end{subfigure}
    \hfill
    \begin{subfigure}[t]{0.49\textwidth}
        \centering
        \includegraphics[width=\linewidth]{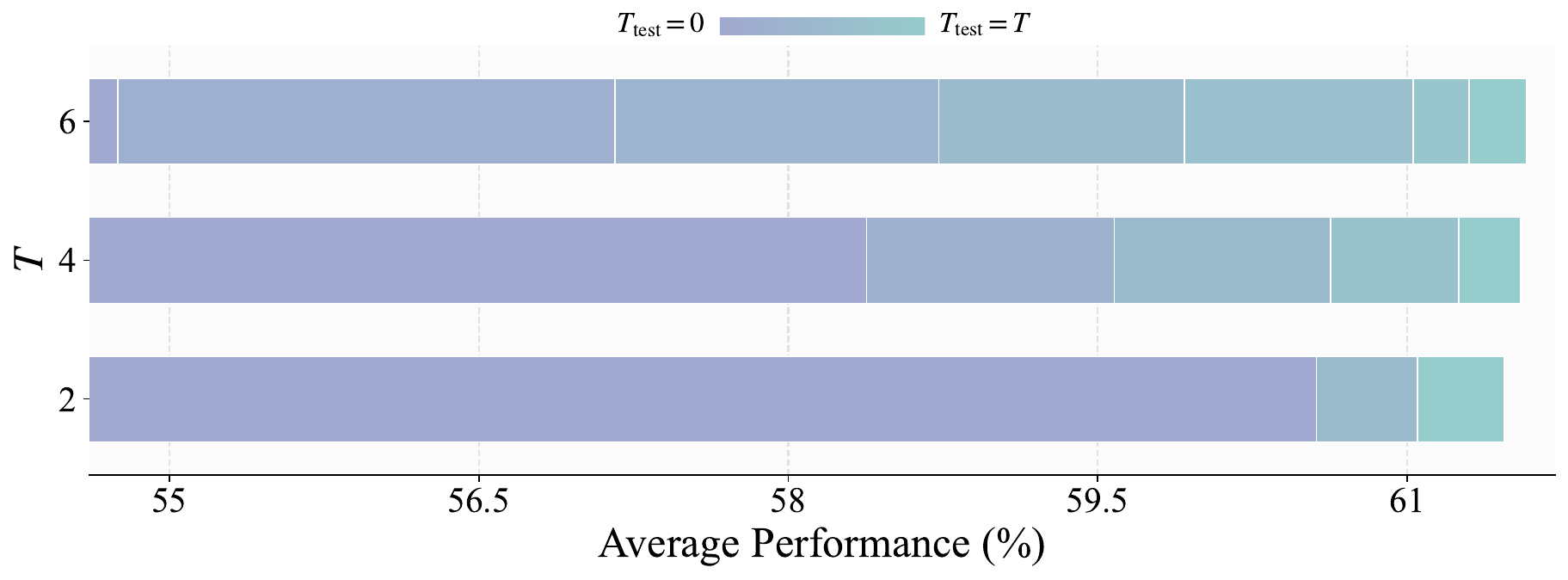}
        \caption{}
        \label{subfigure:test_time_scaling}
    \end{subfigure}
    \vspace{-10pt}
    \caption{\textbf{Robustness and test-time scaling analysis of \ours{} on NLR.} We evaluate how average performance changes with (a) rank $r$, (b) target modules, and (c) inference-time maximum recursion depth $T_{\text{test}}$. Detailed comparisons with additional LoRA variants for (a) and (b) are in~\cref{appendix:analysis}.}
    \label{figure:scaling_analysis}
    \vspace{-10pt}
\end{figure*}

\textbf{Test-Time Scaling.}
We investigate whether \ours{} enables a controllable test-time compute-performance trade-off through adaptive latent recursion. For a model trained with maximum recursion depth $T$, we vary the inference-time maximum recursion depth $T_{\mathrm{test}}\leq T$ while preserving routing and early stopping. \cref{subfigure:test_time_scaling} shows that the average performance consistently improves with increasing $T_{\mathrm{test}}$ across all recursion horizons, and the cumulative gain from $T_{\text{test}}=0$ to $T_{\text{test}}=T$ is larger for checkpoints trained with larger $T$. Consistent with the test-time scaling behavior observed in~\citet{bae2025mor}, these results support that \ours{} can trade additional inference computation for improved downstream performance without retraining.

\begin{wraptable}{r}{0.48\textwidth}
    \centering
    \scriptsize
    \setlength{\tabcolsep}{0.8pt}
    \renewcommand{\arraystretch}{1}
    \caption{\textbf{Computational overhead of LoRA variants on NLR.} We report per-step training time, per-token inference latency, peak GPU memory, and average performance. Detailed profiling settings are provided in~\cref{appendix:analysis}.}
    \label{table:computational_overhead}
    \begin{tabular}{lw{c}{1.05cm}w{c}{1.05cm}w{c}{1.05cm}w{c}{1.05cm}w{c}{0.9cm}}
        \toprule
        \multirow{2}{*}{\raisebox{-0.7ex}{\textbf{Method}}}
        & \multicolumn{2}{c}{\textbf{Training}}
        & \multicolumn{2}{c}{\textbf{Inference}}
        & \multirow{2}{*}{\raisebox{-0.7ex}{\textbf{Avg.}}} \\
        \cmidrule(lr{0pt}){2-3}\cmidrule(lr{3pt}){4-5}
        & \textbf{Time (s)}
        & \textbf{Mem. (GB)}
        & \textbf{Lat. (ms)}
        & \textbf{Mem. (GB)}
        & \\
        \midrule
        LoRA      & 0.55 & 20.76 & 0.18 & 6.09 & 59.21 \\
        TopLoRA   & 0.59 & 21.27 & 0.44 & 6.15 & 60.44 \\
        HydraLoRA & 0.84 & 29.67 & 0.60 & 6.14 & 58.61 \\
        AuroRA    & 0.61 & 21.15 & 1.15 & 6.13 & 59.71 \\
        \ours{}   & 0.62 & 21.12 & 0.69 & 6.13 & 61.21 \\
        \bottomrule
    \end{tabular}
\end{wraptable}

\textbf{Computational Overhead.}
\cref{table:computational_overhead} shows that \ours{} achieves the best overall performance while introducing controlled computational and memory overhead. \ours{} increases per-step training time by about 13\% and peak training memory by less than 2\% over LoRA. At inference, although input-dependent routing prevents \ours{} from being merged into the backbone, its per-token latency falls within the range of the other non-mergeable variants. These results show that \ours{} improves downstream task performance with modest overhead compared with other non-mergeable variants.
%

\subsection{Ablation Studies}
\label{subsection:ablation_studies}
In this subsection, we conduct ablation studies to systematically examine the effects of key design choices in \ours{}. Unless otherwise specified, all experiments use Llama-3.2-3B as the backbone and follow the experimental protocol described in~\cref{subsection:natural_language_reasoning}.

\textbf{Latent Recursion.}
\cref{table:ablation_studies}a compares four latent recursion designs: (i) \textit{no recursion}, which leaves the latent state unchanged; (ii) \textit{fixed depth}, which applies the one-step gated linear transformation as in~\cref{equation:latent_recursion} to all latent states; (iii) \textit{random depth}, which samples recursion depths from a uniform distribution with $T=2$; and (iv) \textit{input-dependent depth}, as defined in~\cref{equation:routing,equation:latent_recursion}. Among the variants designed for the same expected recursion depth, input-dependent depth achieves the highest average performance. Its advantage over uniformly sampled depth suggests that the gains are not explained by depth variation alone and is consistent with a benefit from input-dependent routing.
\begin{table*}[t!]
    \small
    \centering
    \setlength{\fboxsep}{2pt}
    \caption{\textbf{Ablation of \ours{} on NLR.} We investigate different design choices for (a) latent recursion, (b) routing, (c) step conditioning, and evaluate sensitivity to (d) $\eta$. We report accuracy ($\uparrow$) for all other tasks and Pass@1 ($\uparrow$) for HumanEval. The \best{best} results and \colorbox{oursrowcolor!25}{default} designs are highlighted.}
    \setlength{\tabcolsep}{1.7pt}
    \renewcommand{\arraystretch}{\tablestretch}
    \begin{tabular}{
        c
        l
        w{c}{1.2cm}
        w{c}{1.2cm}
        w{c}{1.2cm}
        w{c}{1.2cm}
        w{c}{1.2cm}
        w{c}{1.2cm}
        w{c}{1.2cm}
        |w{c}{1.0cm}
    } 
        \toprule
        & \textbf{Case} 
        & \textbf{ARC-C} 
        & \textbf{BoolQ}
        & \textbf{GSM8K} 
        & \textbf{HS}
        & \textbf{HE}
        & \textbf{MMLU} 
        & \textbf{WG}
        & \textbf{Avg.} 
        \\ 
        \midrule
        \multirow{4}{*}{(a)}
        & No recursion & 49.46 & 88.20 & 34.49 & 78.99 & 23.98 & 55.73 & 83.64 & 59.21 \\
        & Fixed & 54.01 & 88.87 & 39.73 & 78.99 & 23.17 & 55.26 & 83.19 & 60.46 \\
        & Random & 49.83 & 88.15 & 35.41 & 79.01 & 24.59 & 55.60 & 84.08 & 59.53 \\
        & \ourscell Input-dependent
        & \ourscell 54.69
        & \ourscell 89.52
        & \ourscell 40.18
        & \ourscell 79.11
        & \ourscell 25.61
        & \ourscell 55.48
        & \ourscell 83.90
        & \ourscell \best{61.21} \\
        \midrule
        \multirow{4}{*}{(b)}
        & Expert-choice ($\boldsymbol{z}_t$) & 53.70 & 89.57 & 39.80 & 78.95 & 24.59 & 55.46 & 83.87 & 60.85 \\
        & Token-choice ($\boldsymbol{z}_0$) & 54.30 & 89.73 & 40.03 & 79.02 & 23.78 & 55.71 & 83.95 & 60.93 \\
        & Token-choice ($\boldsymbol{x}$) & 54.21 & 89.46 & 40.23 & 79.15 & 23.78 & 55.74 & 84.08 & 60.95 \\
        & \ourscell Threshold-based ($\boldsymbol{z}_t$)
        & \ourscell 54.69
        & \ourscell 89.52
        & \ourscell 40.18
        & \ourscell 79.11
        & \ourscell 25.61
        & \ourscell 55.48
        & \ourscell 83.90
        & \ourscell \best{61.21} \\
        \midrule
        \multirow{5}{*}{(c)}
        & No conditioning              & 54.49 & 89.44 & 40.89 & 79.11 & 24.19 & 55.39 & 83.77 & 61.04 \\
        & \ourscell Fixed sinusoidal
        & \ourscell 54.69
        & \ourscell 89.52
        & \ourscell 40.18
        & \ourscell 79.11
        & \ourscell 25.61
        & \ourscell 55.48
        & \ourscell 83.90
        & \ourscell \best{61.21} \\
        & Learned additive        & 54.32 & 89.34 & 39.60 & 79.04 & 23.78 & 55.79 & 84.29 & 60.88 \\
        & Learned scaling         & 54.44 & 89.29 & 40.49 & 79.07 & 24.80 & 55.44 & 83.82 & 61.05 \\
        & \textcolor{gray!75}{Step-specific transform} & \textcolor{gray!75}{54.35} & \textcolor{gray!75}{89.37} & \textcolor{gray!75}{41.37} & \textcolor{gray!75}{79.05} & \textcolor{gray!75}{25.81} & \textcolor{gray!75}{55.73} & \textcolor{gray!75}{84.06} & \textcolor{gray!75}{61.39} \\
        \midrule
        \multirow{3}{*}{(d)}
        & $\eta=1\times 10^{-2}$ & 54.38 & 89.23 & 39.98 & 79.10 & 25.00 & 55.40 & 83.72 & 60.97 \\
        & \ourscell $\eta=1\times 10^{-3}$
        & \ourscell 54.69
        & \ourscell 89.52
        & \ourscell 40.18
        & \ourscell 79.11
        & \ourscell 25.61
        & \ourscell 55.48
        & \ourscell 83.90
        & \ourscell\best{61.21} \\
        & $\eta=1\times 10^{-4}$ & 54.44 & 89.27 & 40.56 & 79.17 & 25.20 & 55.53 & 83.79 & 61.14 \\
        \bottomrule
    \end{tabular}
    \label{table:ablation_studies}
    \vspace{-15pt}
\end{table*}


\textbf{Routing.}
\cref{table:ablation_studies}b compares different routing strategies: (i) \textit{expert-choice routing}, which selects the highest-scoring tokens at each recursion step based on their current latent states $\boldsymbol{z}_t$; (ii) \textit{token-choice routing from latent states}, which assigns each token a recursion depth in a single decision based on its initial latent state $\boldsymbol{z}_0$; (iii) \textit{token-choice routing from inputs}, which instead predicts the recursion depth directly from $\boldsymbol{x}$; and (iv) \textit{threshold-based routing}, as defined in~\cref{equation:routing}. Threshold-based routing achieves the strongest overall performance. Unlike expert-choice routing, it makes token-wise decisions independently of other tokens, avoiding the information leakage induced by top-$k$ selection in causal models~\citep{bae2025mor}. Unlike one-shot token-choice routing, it re-evaluates whether to continue recursion from the evolving latent state at each step, receiving task-gradient supervision at every executed update. Moreover, the shared router introduces only $r$ trainable parameters, compared with $(T+1)r$ and $(T+1)d_{\mathrm{in}}$ for token-choice routing based on $\boldsymbol{z}_0$ and $\boldsymbol{x}$, respectively. These results indicate that our threshold-based routing can effectively allocate latent computation adaptively while preserving causal compatibility and parameter efficiency.

\textbf{Step Conditioning.}
\cref{table:ablation_studies}c compares different step-conditioning mechanisms: (i) \textit{no conditioning}, which applies the shared latent transformation directly to $\boldsymbol{z}_t$; (ii) \textit{fixed sinusoidal encoding}, which adds $\phi(t)$ to $\boldsymbol{z}_t$; (iii) \textit{learned additive encoding}, which adds a learned step embedding $\boldsymbol{e}_t \in \mathbb{R}^r$; and (iv) \textit{learned scaling}, which applies learned feature-wise scaling. We additionally include \textit{step-specific transform} as a higher-capacity reference, which augments the shared latent transformation $\boldsymbol{W}_{\text{latent}}$ with a zero-initialized, step-specific residual $\Delta\boldsymbol{W}_t \in \mathbb{R}^{r \times r}$. Among the parameter-shared conditioning options, fixed sinusoidal encoding achieves the strongest overall performance, demonstrating the benefit of explicitly conditioning the shared transformation on recursion progress. Its performance is also comparable to the step-specific transform variant while avoiding $Tr^2$ additional parameters per adapted module, yielding a more favorable performance-parameter trade-off.

\textbf{Hyperparameter Sensitivity.}
\cref{table:ablation_studies}d evaluates the sensitivity of \ours{} to the threshold learning rate $\eta$. The results show no significant performance variation across the evaluated scales, with $\eta=1\times10^{-3}$ achieving the best result. This indicates that \ours{} is robust to the choice of $\eta$.

\section{Conclusion}
In this work, we introduce Mixture of Latent Recursions (MoLR), a low-rank adaptation framework that expands latent-space computation without trainable-parameter growth proportional to the maximum recursion depth. MoLR recursively refines the low-rank latent state of each token through a shared transformation, with lightweight threshold-based routing to enable adaptive allocation of latent-space computation across inputs. Extensive experiments across diverse language and vision tasks demonstrate that MoLR achieves the highest performance among competitive LoRA variants under a parameter budget comparable to LoRA. Further analyses reveal token-dependent depth-allocation patterns and a controllable trade-off between inference-time computation and performance within the trained recursion horizon. More broadly, these findings suggest that the low-rank latent space can serve as a reusable computational workspace, pointing to iterative latent refinement as a promising direction for scaling parameter-efficient adaptation.

\subsection*{AI use statement}
In this work, we used generative AI tools for assisting in the writing of
proofs, designing or providing feedback on research methodology or experiments,
and assisting with translation. We have not used generative AI tools for
proposing or refining hypotheses, developing theoretical models or conceptual
frameworks, formulating mathematical claims, providing critical ingredients
for proving mathematical claims, implementing methods, interpreting results, or
supporting qualitative or thematic data analysis, and the remaining tasks with
required disclosure are not applicable to this work.
Additionally, we used generative AI tools for creating or modifying scientific
figures or images, summarizing or analyzing existing literature, brainstorming,
and formatting references. We have reviewed all AI-assisted work. The
AI-assisted proofs and mathematical derivations were verified by the authors.
We take responsibility for the final content of this work, including text,
claims or artifacts produced with the aid of generative AI.

\bibliography{iclr2027_conference}
\bibliographystyle{iclr2027_conference}
\clearpage
\appendix
\section{Experiments}
\label[appendix]{appendix:experiments}

\subsection{Natural Language Understanding}
\label[appendix]{appendix:natural_language_understanding}

\textbf{Datasets.}
Our evaluation is conducted on the GLUE benchmark~\citep{wang2019glue}, a widely adopted suite for measuring general natural language understanding across a broad range of sentence-level and sentence-pair classification tasks. We consider seven GLUE tasks, including CoLA, MNLI, MRPC, QNLI, QQP, SST-2, and STS-B.

\textbf{Training and Evaluation.}
For training, we follow~\citet{wang2024loraga} and employ AdamW with $\beta_1=0.9$, $\beta_2=0.999$, and zero weight decay, together with a cosine learning-rate schedule and a 3\% warmup period. Each task is trained separately using a batch size of 32 and a maximum sequence length of 128. The peak learning rate for the adapters is set to $10^{-3}$ for AdaLoRA and $10^{-4}$ for all other methods to achieve meaningful comparisons. The task-specific classification head is fully fine-tuned at a learning rate of $10^{-4}$. Performance is evaluated using accuracy for all tasks except MRPC, QQP, and STS-B. For MRPC and QQP, we report the mean of accuracy and F1 score, whereas for STS-B, we use the mean of Pearson and Spearman correlation coefficients.

\subsection{Natural Language Reasoning}
\label[appendix]{appendix:natural_language_reasoning}
\textbf{Datasets.}
We evaluate the models across a diverse set of benchmarks, including ARC-Challenge~\citep{clark2018arc}, BoolQ~\citep{clark2019boolq}, HellaSwag~\citep{zellers2019hellaswag}, WinoGrande~\citep{sakaguchi2020winogrande}, MMLU~\citep{hendrycks2021mmlu}, and GSM8K~\citep{cobbe2021gsm8k}. For code generation, models are trained on CodeAlpaca-20K~\citep{chaudhary2023codealpaca} and evaluated on HumanEval~\citep{chen2021humaneval}. These benchmarks cover a broad range of capabilities, including commonsense reasoning, reading comprehension, general knowledge understanding, mathematical reasoning, and code generation.

\textbf{Training and Evaluation.}
We separately fine-tune on each benchmark with AdamW, using $\beta_1=0.9$, $\beta_2=0.999$, no weight decay, and a linear learning-rate schedule with 3\% warmup. We use a batch size of 16 throughout. AdaLoRA is trained with a peak learning rate of $5\times10^{-4}$, whereas all other approaches use $5\times10^{-5}$ for meaningful comparisons. The input length is capped at 128 tokens for ARC-Challenge, HellaSwag, and MMLU, 256 tokens for BoolQ and WinoGrande, and 512 tokens for GSM8K and CodeAlpaca-20K. Evaluation is performed with \texttt{lm-evaluation-harness}~\citep{eval-harness}. We report length-normalized accuracy on ARC-Challenge and HellaSwag, exact-match accuracy on GSM8K, pass@1 on HumanEval, and standard accuracy on the remaining benchmarks.

\subsection{Image Classification}
\label[appendix]{appendix:image_classification}
\textbf{Datasets.}
We evaluate the models on a diverse set of image classification benchmarks, including Stanford Cars~\citep{krause2013cars}, DTD~\citep{cimpoi2014dtd}, EuroSAT~\citep{helber2019eurosat}, GTSRB~\citep{stallkamp2012gtsrb}, RESISC45~\citep{cheng2017resisc45}, SUN397~\citep{xiao2010sun}, and SVHN~\citep{netzer2011svhn}. These datasets cover a broad range of visual recognition settings.

\textbf{Training and Evaluation.}
We fine-tune on individual tasks using AdamW with $\beta_1=0.9$, $\beta_2=0.999$, zero weight decay, and a cosine learning-rate schedule with 3\% warmup. Adapters are applied only to the linear modules of the vision transformer. We use a batch size of 64 and set the peak learning rate to $10^{-3}$ for AdaLoRA and $10^{-4}$ for all other methods. Predictions are obtained using the prompt template ``\texttt{a photo of a {class}}''. 

\subsection{Analysis}
\label[appendix]{appendix:analysis}

\textbf{Across-Seed Variability.}
\cref{table:natural_language_understanding_standard_deviation,table:natural_language_reasoning_standard_deviation,table:image_classification_standard_deviation} report the standard deviations corresponding to the quantitative comparisons on NLU, NLR, and IC, respectively. The observed variation is consistently small relative to the performance improvements delivered by \ours{}. This suggests that the gains of \ours{} are stable across repeated runs and are unlikely to arise from experimental variability.
\begin{table*}[t!]
    \small
    \centering
    \setlength{\fboxsep}{2pt}
    \caption{\textbf{Standard deviations for the quantitative comparison of LoRA variants on NLU.} We report the standard deviations corresponding to the results in \cref{table:natural_language_understanding}.}
    \setlength{\tabcolsep}{4.9pt}
    \renewcommand{\arraystretch}{\tablestretch}
    \begin{tabular}{
        c
        l
        w{c}{1.0cm}
        w{c}{1.0cm}
        w{c}{1.0cm}
        w{c}{1.0cm}
        w{c}{1.0cm}
        w{c}{1.0cm}
        w{c}{1.0cm}
        |w{c}{1.0cm}
    }
        \toprule
        \textbf{Model}
        & \textbf{Method}
        & \textbf{CoLA}
        & \textbf{MNLI}
        & \textbf{MRPC}
        & \textbf{QNLI}
        & \textbf{QQP}
        & \textbf{SST-2}
        & \textbf{STSB}
        & \textbf{Avg.}
        \\
        \midrule
        \multirow{8}{*}{\rotatebox{90}{RoBERTa-B}}
        & LoRA      & 0.58 & 0.12 & 0.39 & 0.22 & 0.05 & 0.40 & 0.48 & 0.08 \\
        & DoRA      & 0.25 & 0.81 & 0.16 & 0.20 & 0.09 & 0.29 & 0.55 & 0.16 \\
        & AdaLoRA   & 0.59 & 0.17 & 0.18 & 0.32 & 0.01 & 0.31 & 0.13 & 0.12 \\
        & TopLoRA   & 0.20 & 0.25 & 0.87 & 0.34 & 0.06 & 0.43 & 0.02 & 0.07 \\
        & HydraLoRA & 0.72 & 0.06 & 0.85 & 0.34 & 0.09 & 0.13 & 0.12 & 0.13 \\
        & MoSLoRA   & 0.64 & 0.20 & 0.33 & 0.12 & 0.09 & 0.40 & 0.80 & 0.19 \\
        & AuroRA    & 0.25 & 0.12 & 0.28 & 0.14 & 0.14 & 0.50 & 0.53 & 0.12 \\
        & \ours{}   & 0.33 & 0.12 & 0.47 & 0.33 & 0.12 & 0.63 & 0.53 & 0.05 \\
        \midrule
        \multirow{8}{*}{\rotatebox{90}{DeBERTa-v3-L}}
        & LoRA      & 0.31 & 0.14 & 0.10 & 0.05 & 0.12 & 0.41 & 0.49 & 0.10 \\
        & DoRA      & 0.22 & 0.13 & 0.10 & 0.14 & 0.04 & 0.17 & 0.72 & 0.12 \\
        & AdaLoRA   & 0.06 & 0.09 & 0.84 & 0.05 & 0.10 & 0.17 & 0.49 & 0.15 \\
        & TopLoRA   & 0.33 & 0.15 & 0.97 & 0.45 & 0.06 & 0.35 & 0.58 & 0.14 \\
        & HydraLoRA & 0.72 & 0.17 & 0.87 & 0.15 & 0.01 & 0.07 & 0.72 & 0.27 \\
        & MoSLoRA   & 0.44 & 0.22 & 0.72 & 0.05 & 0.05 & 0.23 & 0.68 & 0.17 \\
        & AuroRA    & 0.33 & 0.04 & 0.17 & 0.14 & 0.04 & 0.18 & 0.77 & 0.07 \\
        & \ours{}   & 0.40 & 0.03 & 0.08 & 0.09 & 0.06 & 0.17 & 0.37 & 0.07 \\
        \bottomrule
    \end{tabular}
    \label{table:natural_language_understanding_standard_deviation}
    \vspace{-5pt}
\end{table*}


\textbf{Robustness.}
\cref{table:rank_capacity_sweep,table:target_module_sweep} extend the robustness analysis in~\cref{figure:scaling_analysis} to representative LoRA variants. \ours{} achieves the highest average performance across all evaluated ranks and target-module configurations, demonstrating that its gains persist across a broad range of adaptation budgets. Notably, \ours{} at lower ranks outperforms competing variants at higher ranks, including methods that already incorporate explicit latent-space computation. This indicates that adaptively refining a compact latent representation can yield greater expressivity than simply increasing its dimensionality, and further suggests that scaling latent-space computation constitutes a complementary axis for increasing adaptation capacity beyond the conventional strategy of expanding the latent rank.

\textbf{Computational Overhead.}
We evaluate computational efficiency on a single NVIDIA RTX A6000 using Llama-3.2-3B as the backbone and an adapter rank of $r=8$, consistent with the setup in~\cref{subsection:natural_language_reasoning}. Training efficiency is assessed with a batch size of 16 and a sequence length of 128 by recording the average step latency and peak CUDA memory across 50 synchronized optimization steps following 10 warm-up steps. For inference, we similarly measure per-token latency and peak CUDA memory over 50 synchronized next-token forward passes after 10 warm-up batches. Whenever possible, adapter parameters are merged into the backbone weights before inference.

\begin{wraptable}{r}{0.48\textwidth}
    \centering
    \scriptsize
    \setlength{\tabcolsep}{0.8pt}
    \vspace{-\intextsep}
    \renewcommand{\arraystretch}{1}
    \caption{\textbf{Compatibility with existing LoRA variants.} We report average performance ($\uparrow$) on diverse downstream tasks. \#Param. denotes the number of trainable parameters. * shows that the variant is augmented with adaptive latent recursion. \textbf{Better} results are highlighted.}
    \vspace{-7pt}
    \label{table:compatibility}
    \begin{tabular}{l*{3}{w{c}{1.0cm}w{c}{0.7cm}}}
        \toprule
        \multirow{2}{*}{\raisebox{-0.7ex}{\textbf{Method}}}
        & \multicolumn{2}{c}{\textbf{NLU}}
        & \multicolumn{2}{c}{\textbf{NLR}}
        & \multicolumn{2}{c}{\textbf{CV}} \\
        \cmidrule(lr{0pt}){2-3}\cmidrule(lr{3pt}){4-5}\cmidrule(lr{3pt}){6-7}
        & \textbf{\#Param.} & \textbf{Avg.}
        & \textbf{\#Param.} & \textbf{Avg.}
        & \textbf{\#Param.} & \textbf{Avg.} \\
        \midrule
        VeRA  & 0.10M & 77.00 & 0.82M & 52.66 & 0.10M & 56.08 \\
        VeRA* & 0.20M & \textbf{81.02} & 0.99M & \textbf{54.70} & 0.20M & \textbf{67.14} \\
        \midrule
        LoRA-FA  & 0.66M & 85.08 & 6.19M & 59.05 & 0.66M & 85.29 \\
        LoRA-FA* & 0.67M & \textbf{85.37} & 6.21M & \textbf{59.29} & 0.67M & \textbf{85.46} \\
        \midrule
        AuroRA  & 1.34M & 84.94 & 12.18M & 59.71 & 1.34M & 82.26 \\
        AuroRA* & 1.34M & \textbf{85.88} & 12.18M & \textbf{60.18} & 1.34M & \textbf{82.86} \\
        \bottomrule
    \end{tabular}
    \vspace{-\intextsep}
\end{wraptable}

\textbf{Compatibility.}
We evaluate the generalizability of adaptive latent recursion by integrating it into three existing LoRA variants: VeRA~\citep{kopiczko2024vera}, LoRA-FA~\citep{zhang2023lorafa}, and AuroRA~\citep{dong2025aurora}. These methods span diverse latent-space constructions: VeRA employs frozen shared matrices, LoRA-FA freezes the down-projection, and AuroRA learns nonlinear latent transformations without frozen projections. \cref{table:compatibility} shows that adaptive latent recursion consistently improves all three variants across all tasks. These gains demonstrate that adaptive latent recursion generalizes across different LoRA formulations, enhancing adaptation expressiveness while preserving parameter efficiency.
\begin{table*}[t!]
    \small
    \centering
    \setlength{\fboxsep}{2pt}
    \caption{\textbf{Standard deviations for the quantitative comparison of LoRA variants on NLR.} We report the standard deviations corresponding to the results in \cref{table:natural_language_reasoning}.}
    \setlength{\tabcolsep}{4.9987pt}
    \renewcommand{\arraystretch}{\tablestretch}
    \begin{tabular}{
        c
        l
        w{c}{1.0cm}
        w{c}{1.0cm}
        w{c}{1.0cm}
        w{c}{1.0cm}
        w{c}{1.0cm}
        w{c}{1.0cm}
        w{c}{1.0cm}
        |w{c}{1.0cm}
    }
        \toprule
        \textbf{Model}
        & \textbf{Method}
        & \textbf{ARC-C}
        & \textbf{BoolQ}
        & \textbf{GSM8K}
        & \textbf{HS}
        & \textbf{HE}
        & \textbf{MMLU}
        & \textbf{WG}
        & \textbf{Avg.}
        \\
        \midrule
        \multirow{8}{*}{\rotatebox{90}{LLaMA-3.2-3B}}
        & LoRA      & 0.13 & 0.25 & 0.33 & 0.12 & 1.27 & 0.31 & 0.32 & 0.23 \\
        & DoRA      & 0.22 & 0.42 & 0.40 & 0.06 & 1.96 & 0.24 & 0.59 & 0.21 \\
        & AdaLoRA   & 0.20 & 0.22 & 0.31 & 0.05 & 0.35 & 0.22 & 0.37 & 0.02 \\
        & TopLoRA   & 0.40 & 0.23 & 0.31 & 0.05 & 0.35 & 0.15 & 0.71 & 0.12 \\
        & HydraLoRA & 0.55 & 0.36 & 1.03 & 0.09 & 0.70 & 0.25 & 0.44 & 0.17 \\
        & MoSLoRA   & 0.24 & 0.22 & 0.40 & 0.13 & 0.93 & 0.23 & 0.09 & 0.14 \\
        & AuroRA    & 0.46 & 0.12 & 1.16 & 0.09 & 1.27 & 0.39 & 0.40 & 0.13 \\
        & \ours{}   & 0.22 & 0.29 & 0.35 & 0.12 & 1.83 & 0.34 & 0.72 & 0.26 \\
        \midrule
        \multirow{8}{*}{\rotatebox{90}{Qwen-3-8B}}
        & LoRA      & 1.69 & 0.22 & 0.50 & 0.18 & 2.14 & 0.15 & 0.39 & 0.10 \\
        & DoRA      & 1.83 & 0.18 & 0.33 & 0.01 & 2.66 & 0.13 & 0.55 & 0.32 \\
        & AdaLoRA   & 1.72 & 0.17 & 0.31 & 0.10 & 2.19 & 0.39 & 0.25 & 0.41 \\
        & TopLoRA   & 1.54 & 0.14 & 0.42 & 0.13 & 1.06 & 0.11 & 0.59 & 0.26 \\
        & HydraLoRA & 1.74 & 0.18 & 0.62 & 0.06 & 1.96 & 0.14 & 0.09 & 0.09 \\
        & MoSLoRA   & 1.87 & 0.07 & 0.32 & 0.12 & 3.39 & 0.16 & 0.18 & 0.49 \\
        & AuroRA    & 1.43 & 0.10 & 0.53 & 0.05 & 0.35 & 0.17 & 0.18 & 0.10 \\
        & \ours{}   & 0.33 & 0.05 & 0.17 & 0.09 & 1.17 & 0.23 & 0.27 & 0.43 \\
        \bottomrule
    \end{tabular}
    \label{table:natural_language_reasoning_standard_deviation}
    \vspace{-5pt}
\end{table*}

\begin{table*}[t!]
    \small
    \centering
    \setlength{\fboxsep}{2pt}
    \caption{\textbf{Standard deviations for the quantitative comparison of LoRA variants on IC.} We report the standard deviations corresponding to the results in \cref{table:image_classification}.}
    \setlength{\tabcolsep}{3.8pt}
    \renewcommand{\arraystretch}{\tablestretch}
    \begin{tabular}{
        c
        l
        w{c}{1.1cm}
        w{c}{1.1cm}
        w{c}{1.1cm}
        w{c}{1.1cm}
        w{c}{1.2cm}
        w{c}{1.1cm}
        w{c}{1.1cm}
        |w{c}{1.0cm}
    }
        \toprule
        \textbf{Model}
        & \textbf{Method}
        & \textbf{Cars}
        & \textbf{DTD}
        & \textbf{ESAT}
        & \textbf{GTSRB}
        & \textbf{RE45}
        & \textbf{SUN397}
        & \textbf{SVHN}
        & \textbf{Avg.}
        \\
        \midrule
        \multirow{8}{*}{\rotatebox{90}{CLIP-ViT-B}}
        & LoRA      & 0.28 & 0.31 & 0.07 & 0.25 & 0.26 & 0.05 & 0.09 & 0.08 \\
        & DoRA      & 0.27 & 0.20 & 0.13 & 0.24 & 0.25 & 0.07 & 0.06 & 0.13 \\
        & AdaLoRA   & 0.15 & 0.16 & 0.09 & 0.03 & 0.15 & 0.15 & 0.07 & 0.01 \\
        & TopLoRA   & 0.38 & 0.52 & 0.10 & 0.10 & 0.08 & 0.17 & 0.09 & 0.15 \\
        & HydraLoRA & 0.27 & 0.58 & 0.05 & 0.06 & 0.01 & 0.10 & 0.08 & 0.07 \\
        & MoSLoRA   & 0.35 & 0.23 & 0.10 & 0.12 & 0.23 & 0.12 & 0.05 & 0.04 \\
        & AuroRA    & 0.26 & 0.76 & 0.14 & 0.30 & 0.24 & 0.19 & 0.04 & 0.14 \\
        & \ours{}   & 0.09 & 0.27 & 0.07 & 0.08 & 0.25 & 0.05 & 0.05 & 0.05 \\
        \midrule
        \multirow{8}{*}{\rotatebox{90}{SigLIP-ViT-L}}
        & LoRA      & 0.04 & 0.13 & 0.22 & 0.07 & 0.03 & 0.13 & 0.02 & 0.01 \\
        & DoRA      & 0.05 & 0.03 & 0.10 & 0.11 & 0.06 & 0.05 & 0.06 & 0.03 \\
        & AdaLoRA   & 0.09 & 0.06 & 0.17 & 0.10 & 0.12 & 0.04 & 0.09 & 0.02 \\
        & TopLoRA   & 0.04 & 0.25 & 0.05 & 0.09 & 0.08 & 0.05 & 0.02 & 0.06 \\
        & HydraLoRA & 0.06 & 0.03 & 0.08 & 0.04 & 0.15 & 0.05 & 0.03 & 0.01 \\
        & MoSLoRA   & 0.07 & 0.06 & 0.11 & 0.02 & 0.19 & 0.05 & 0.05 & 0.01 \\
        & AuroRA    & 0.08 & 0.19 & 0.06 & 0.03 & 0.12 & 0.10 & 0.05 & 0.03 \\
        & \ours{}   & 0.06 & 0.18 & 0.10 & 0.08 & 0.12 & 0.14 & 0.06 & 0.08 \\
        \bottomrule
    \end{tabular}
    \label{table:image_classification_standard_deviation}
    \vspace{-5pt}
\end{table*}

\begin{table*}[t!]
    \small
    \centering
    \caption{\textbf{Quantitative comparison of LoRA variants under different rank capacities on NLR.} We sweep rank capacities and report accuracy ($\uparrow$) for all other tasks and Pass@1 ($\uparrow$) for HumanEval. The \textbf{best} results are highlighted. HS = HellaSwag, HE = HumanEval, WG = WinoGrande.}
    \setlength{\tabcolsep}{1.6pt}
    \renewcommand{\arraystretch}{\tablestretch}
    \begin{tabular}{
        c
        l
        w{c}{1.2cm}
        w{c}{1.2cm}
        w{c}{1.2cm}
        w{c}{1.2cm}
        w{c}{1.2cm}
        w{c}{1.2cm}
        w{c}{1.2cm}
        |w{c}{1.0cm}
    }
        \toprule
        \textbf{Rank Capacity}
        & \textbf{Method}
        & \textbf{ARC-C}
        & \textbf{BoolQ}
        & \textbf{GSM8K}
        & \textbf{HS}
        & \textbf{HE}
        & \textbf{MMLU}
        & \textbf{WG}
        & \textbf{Avg.}
        \\
        \midrule

        \multirow{5}{*}{$r=8$}
        & LoRA    & 49.46 & 88.20 & 34.49 & 78.99 & 23.98 & 55.73 & 83.64 & 59.21 \\
        & TopLoRA & 50.97 & 88.84 & 39.83 & 79.42 & 24.19 & 56.25 & 83.58 & 60.44 \\
        & MoSLoRA & 52.16 & 89.33 & 38.06 & 78.78 & 22.36 & 54.46 & 83.71 & 59.84 \\
        & AuroRA  & 52.64 & 88.91 & 35.51 & 79.05 & 24.80 & 53.34 & 83.74 & 59.71 \\
        & \ourscell\ours{} & \ourscell 54.69 & \ourscell 89.52 & \ourscell 40.18 & \ourscell 79.11 & \ourscell 25.61 & \ourscell 55.48 & \ourscell 83.90 & \ourscell\best{61.21} \\
        \midrule

        \multirow{5}{*}{$r=16$}
        & LoRA    & 50.99 & 88.70 & 35.66 & 79.16 & 24.80 & 55.87 & 84.16 & 59.91 \\
        & TopLoRA & 52.42 & 89.36 & 41.72 & 79.54 & 23.37 & 56.04 & 83.14 & 60.80 \\
        & MoSLoRA & 53.10 & 89.50 & 39.47 & 78.97 & 25.00 & 55.38 & 83.53 & 60.71 \\
        & AuroRA  & 52.30 & 88.79 & 37.55 & 79.10 & 22.97 & 52.85 & 84.03 & 59.66 \\
        & \ourscell\ours{} & \ourscell 54.24 & \ourscell 89.52 & \ourscell 40.66 & \ourscell 79.32 & \ourscell 25.61 & \ourscell 55.77 & \ourscell 83.98 & \ourscell\best{61.30} \\
        \midrule

        \multirow{5}{*}{$r=32$}
        & LoRA    & 52.22 & 89.00 & 37.30 & 79.24 & 24.19 & 56.25 & 83.90 & 60.30 \\
        & TopLoRA & 53.70 & 89.33 & 41.92 & 79.36 & 22.97 & 55.99 & 83.48 & 60.96 \\
        & MoSLoRA & 53.24 & 89.37 & 38.99 & 79.23 & 24.59 & 55.85 & 84.53 & 60.83 \\
        & AuroRA  & 46.64 & 85.49 & 37.55 & 78.79 & 24.80 & 50.27 & 83.29 & 58.12 \\
        & \ourscell\ours{} & \ourscell 54.61 & \ourscell 89.43 & \ourscell 41.47 & \ourscell 79.31 & \ourscell 25.00 & \ourscell 56.11 & \ourscell 84.40 & \ourscell\best{61.47} \\
        \bottomrule
    \end{tabular}
    \label{table:rank_capacity_sweep}
    \vspace{-5pt}
\end{table*}

\begin{table*}[t!]
    \small
    \centering
    \caption{\textbf{Quantitative comparison of LoRA-based methods under different target-module sets on NLR.} We vary the target-module set and report accuracy ($\uparrow$) for all other tasks and Pass@1 ($\uparrow$) for HumanEval. \texttt{attn} includes $\{\boldsymbol{W}_{\text{Q}}, \boldsymbol{W}_{\text{K}}, \boldsymbol{W}_{\text{V}}, \boldsymbol{W}_{\text{O}}\}$, \texttt{mlp} includes $\{\boldsymbol{W}_{\text{Gate}}, \boldsymbol{W}_{\text{Up}}, \boldsymbol{W}_{\text{Down}}\}$, and \texttt{all} includes their union. The \textbf{best} results are highlighted. HS = HellaSwag, HE = HumanEval, WG = WinoGrande.}
    \setlength{\tabcolsep}{1.6pt}
    \renewcommand{\arraystretch}{\tablestretch}
    \begin{tabular}{
        c
        l
        w{c}{1.2cm}
        w{c}{1.2cm}
        w{c}{1.2cm}
        w{c}{1.2cm}
        w{c}{1.2cm}
        w{c}{1.2cm}
        w{c}{1.2cm}
        |w{c}{1.0cm}
    }
        \toprule
        \textbf{Target Module}
        & \textbf{Method}
        & \textbf{ARC-C}
        & \textbf{BoolQ}
        & \textbf{GSM8K}
        & \textbf{HS}
        & \textbf{HE}
        & \textbf{MMLU}
        & \textbf{WG}
        & \textbf{Avg.}
        \\
        \midrule
        \multirow{5}{*}{\texttt{attn}}
        & LoRA    & 47.52 & 87.24 & 32.80 & 78.51 & 25.81 & 55.08 & 82.32 & 58.47 \\
        & TopLoRA & 49.63 & 88.43 & 36.11 & 79.09 & 23.58 & 55.22 & 82.58 & 59.24 \\
        & MoSLoRA & 49.89 & 88.30 & 34.42 & 78.56 & 23.37 & 54.46 & 81.53 & 58.65 \\
        & AuroRA  & 50.17 & 88.12 & 33.53 & 78.65 & 25.61 & 53.32 & 82.45 & 58.84 \\
        & \ourscell\ours{} & \ourscell 51.93 & \ourscell 88.99 & \ourscell 36.74 & \ourscell 78.68 & \ourscell 23.98 & \ourscell 54.82 & \ourscell 82.24 & \ourscell\best{59.63} \\
        \midrule
        \multirow{5}{*}{\texttt{mlp}}
        & LoRA    & 47.84 & 87.63 & 33.84 & 78.92 & 23.78 & 55.46 & 83.39 & 58.70 \\
        & TopLoRA & 50.71 & 88.61 & 38.16 & 79.31 & 23.78 & 55.93 & 83.92 & 60.06 \\
        & MoSLoRA & 51.25 & 88.47 & 35.73 & 78.73 & 24.39 & 55.26 & 83.74 & 59.65 \\
        & AuroRA  & 51.37 & 88.23 & 35.00 & 78.93 & 23.37 & 54.00 & 82.93 & 59.12 \\
        & \ourscell\ours{} & \ourscell 52.93 & \ourscell 88.91 & \ourscell 38.44 & \ourscell 78.99 & \ourscell 24.59 & \ourscell 55.65 & \ourscell 83.66 & \ourscell\best{60.45} \\
        \midrule
        \multirow{5}{*}{\texttt{all}}
        & LoRA    & 49.46 & 88.20 & 34.49 & 78.99 & 23.98 & 55.73 & 83.64 & 59.21 \\
        & TopLoRA & 50.97 & 88.84 & 39.83 & 79.42 & 24.19 & 56.25 & 83.58 & 60.44 \\
        & MoSLoRA & 52.16 & 89.33 & 38.06 & 78.78 & 22.36 & 54.46 & 83.71 & 59.84 \\
        & AuroRA  & 52.64 & 88.91 & 35.51 & 79.05 & 24.80 & 53.34 & 83.74 & 59.71 \\
        & \ourscell\ours{} & \ourscell 54.69 & \ourscell 89.52 & \ourscell 40.18 & \ourscell 79.11 & \ourscell 25.61 & \ourscell 55.48 & \ourscell 83.90 & \ourscell\best{61.21} \\
        \bottomrule
    \end{tabular}
    \label{table:target_module_sweep}
    \vspace{-5pt}
\end{table*}

\begin{figure}[t]
    \centering
    \includegraphics[width=0.75\textwidth]{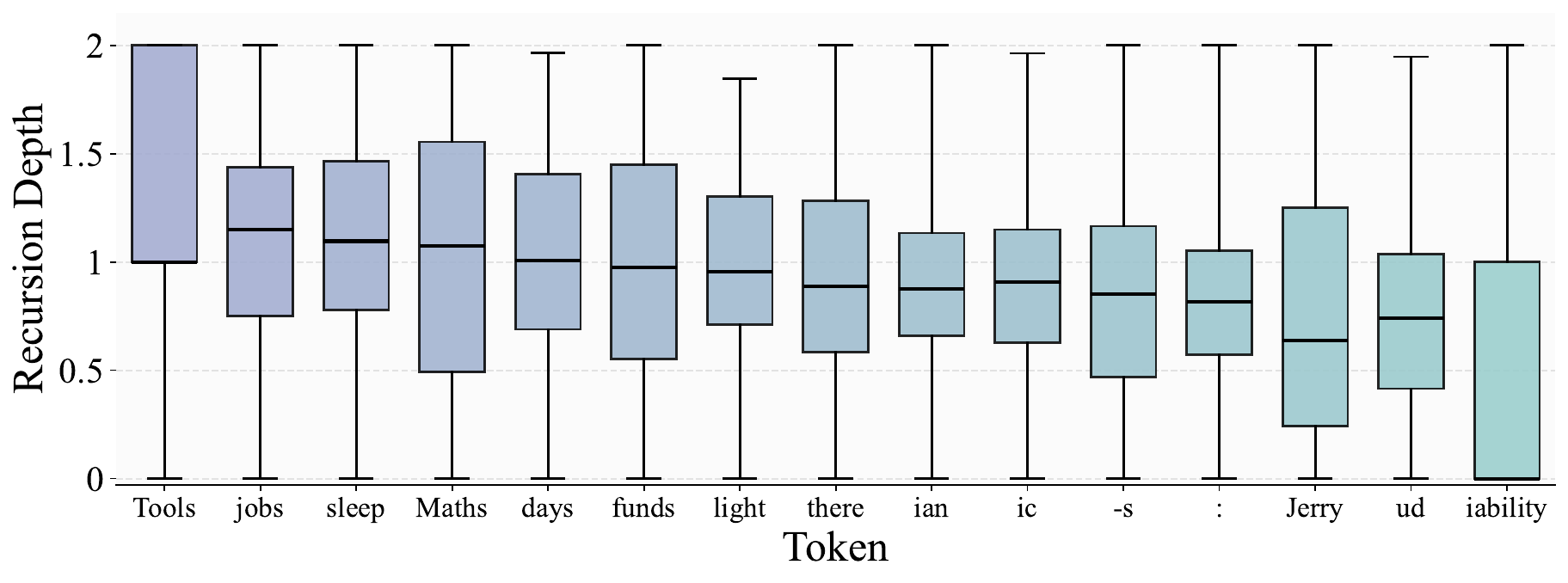}
    \vspace{-10pt}
    \caption{\textbf{Distribution of recursion depth across tokens on MMLU.} Box plots report the median, interquartile range, and minimum/maximum values.}
    \label{figure:recursion_depth_per_token_mmlu}
\end{figure}

\textbf{Recursion Depth Allocation.}
We analyze the distribution of recursion depths for tokens sampled uniformly from the vocabulary based on recursion utilization in \cref{figure:recursion_depth_per_token_mmlu}. Each box summarizes the average recursion depths over token occurrences within all adapted modules, with the box indicating the interquartile range, the central line denoting the median, and the whiskers representing the minimum and maximum. We observe that semantically complete words (e.g., ``Tools", ``jobs", ``sleep") generally receive greater recursion depths than subword fragments (e.g., ``ic", ``-s", ``ud"). These findings support that MoLR learns heterogeneous recursion allocations implicitly from task loss.

\section{Discussion}
\label[appendix]{appendix:discussion}
In this section, we discuss \ours{} from several perspectives, including its empirical behavior and broader implications, current limitations, and promising directions for future work.

\textbf{Exact Expansion of Adaptive Latent Recursion}
\label[appendix]{appendix:exact_latent_recursion}
For an input that executes $D(\boldsymbol{x})$ recursion steps, expanding~\cref{equation:latent_recursion} gives
\begin{equation}
\boldsymbol{z}_{D(\boldsymbol{x})}
=\left[\prod_{t=0}^{D(\boldsymbol{x})-1}\left(\boldsymbol{I}_r+s_t\boldsymbol{W}_{\text{latent}}\right)\right]\boldsymbol{z}_0
+\sum_{k=0}^{D(\boldsymbol{x})-1}\left[\prod_{j=k+1}^{D(\boldsymbol{x})-1}\left(\boldsymbol{I}_r+s_j\boldsymbol{W}_{\text{latent}}\right)\right]s_k\boldsymbol{W}_{\text{latent}}\phi(k).
\end{equation}
The exact expansion shows that a fixed step encoding provides a stage signal whose effect is shaped by subsequent shared updates and state-dependent routing. This may explain why learning separate step encodings is unnecessary in this setting, as the shared transformation and router can learn how to use a deterministic signal. Consistent with this interpretation, fixed sinusoidal conditioning has the highest average result among the parameter-shared conditioning variants as shown in~\cref{table:ablation_studies}.

\textbf{Threshold-based Routing.}
As shown in~\cref{equation:latent_recursion}, continuation is a discrete decision, and we do not propagate surrogate gradients through its binary indicator. For methodological simplicity, we separate task-driven router learning from load balancing. Specifically, the router receives task-loss gradients through the continuous score $s_t$ that scales executed updates, and the threshold $b_t$ is adjusted from observed continuation rates by~\cref{equation:load_balancing}. The threshold determines whether an update executes but does not scale its magnitude once selected, so balancing adds no auxiliary gradient to the task objective. Whether surrogate gradients through the discrete continuation decision are necessary or beneficial, and whether they would blur the separation between task-driven router learning and load-driven threshold adjustment, remain open questions for future work.

\textbf{Input-Dependent Computation Allocation.}
\ours{} allocates latent-space computation according to the input, allowing different inputs to execute different numbers of recursion steps. This complements rank-allocation methods, which adjust the rank assigned to adapted weight matrices according to their importance~\citep{zhang2023increlora,zhang2023adalora,ding2023sora}. While these methods redistribute adaptation capacity across modules, increasing the total available rank still requires additional trainable rank components. In contrast, \ours{} varies the computation applied to each input through a shared latent transformation, without trainable-parameter growth proportional to recursion depth.

\textbf{Test-Time Scaling.}
For a fixed \ours{} checkpoint trained with maximum recursion depth $T$, varying $T_{\text{test}}\leq T$ adjusts the inference-time adapter compute budget without requiring additional training. As shown in~\cref{subfigure:test_time_scaling}, performance improves as $T_{\text{test}}$ increases within each evaluated training horizon. This equips a frozen pre-trained backbone with a bounded test-time compute knob through a lightweight adapter, without requiring the recurrent backbone itself to be pre-trained from scratch~\citep{geiping2025recurrent,jeddi2026loopformer,yu2026tloopformer}. Consequently, a single adapted checkpoint can operate under different compute budgets, using a smaller depth cap when latency is constrained and allowing additional latent refinement when more computation is available. Token-level halting further enables this budget to be allocated non-uniformly across inputs, potentially concentrating computation where additional refinement is most beneficial. More broadly, this mechanism may make adaptive inference-time computation accessible to existing language and vision models through parameter-efficient adaptation. It also provides a potentially complementary scaling axis to chain-of-thought prompting~\citep{wei2022chain}, where explicit multi-round reasoning can extend the sequence of steps and adaptive latent recursion can vary the computation applied within each step. We leave these open questions for future work.

\textbf{Computational Overhead.}
Due to input-dependent routing and recursion, \ours{} cannot be merged into the pre-trained weights. As shown in~\cref{table:computational_overhead}, it achieves the highest average performance with modest training overhead, while maintaining inference latency and memory comparable to other non-mergeable LoRA variants. Further efficiency improvements through algorithmic and hardware optimizations remain future work.

\end{document}